\RequirePackage{fix-cm}
\documentclass[smallcondensed]{svjour3}     % onecolumn (ditto)
\smartqed  % flush right qed marks, e.g. at end of proof
\usepackage{graphicx}
\usepackage{mathptmx,epsfig,latexsym,epstopdf, subfig}
\usepackage[numbers]{natbib}
\DeclareGraphicsExtensions{.eps}
\journalname{Arxiv}
\begin{document}

\title{A Cognitive Architecture Based on a Learning Classifier System with Spiking Classifiers%\thanks{Grants or other notes
}

\titlerunning{A Cognitive Architecture Based on an Spiking LCS}        % if too long for running head

\author{David Howard        \and
        Larry Bull \and
        Pier-Luca Lanzi %etc.
}


\institute{G. Howard \at
                Autonomous Systems Program, Queensland Centre for Advanced Technology, Australia\\
              \email{david.howard@csiro.au}
           \and
          L. Bull \at
               Faculty of Environment and Technology, University of the West of England, UK
           \and
           P.-L. Lanzi \at
              Dipartimento di Elettronica e Informazione, Politecnico di Milano, Italy
}


\maketitle

\begin{abstract}
Learning Classifier Systems (LCS) are population-based reinforcement learners that were originally designed to model various cognitive phenomena.  This paper presents an explicitly cognitive LCS by using spiking neural networks as classifiers, providing each classifier with a measure of temporal dynamism.  We employ a constructivist model of growth of both neurons and synaptic connections, which permits a Genetic Algorithm (GA) to automatically evolve sufficiently-complex neural structures.  The spiking classifiers are coupled with a temporally-sensitive reinforcement learning algorithm, which allows the system to perform temporal state decomposition by appropriately rewarding  ``macro-actions'', created by chaining together multiple atomic actions.  The combination of temporal reinforcement learning and neural information processing is shown to outperform benchmark neural classifier systems, and successfully solve a robotic navigation task.

\keywords{Learning Classifier Systems \and Spiking Neural Networks \and Self-adaptation \and Semi-MDP}
\end{abstract}

%%%%%%%%%%%%%%%%%%%%%
%%%%%%%%%%%%%%%%%%%%% INTRODUCTION
%%%%%%%%%%%%%%%%%%%%%

\section{Introduction}
\label{intro}

Learning Classifier Systems (LCS) \citep{Holland76} --- for example the first LCS,  CS1 \cite{Holland1978} --- were originally devised to investigate cognitive systems that model the interation of a Genetic Algorithm (GA) \cite{Holland75} and Reinforcement Learning (RL) \cite{sutton-scc}.  An LCS involves a population of classifiers whose condition determines which subspaces of the state space they are active in,  and an action which is advocated in those subspaces.  The GA is used to generate useful condition, action pairings (i.e., when a classifier is active and what it does when it is active), and RL is used to appropriately reward the results of taking an action in a given state.  In this way, an LCS can generate optimal action selection policies in an environment.  An LCS therefore consists of three main components: a GA, an RL scheme, and a population of classifiers.

When viewing an LCS as a cognitive architecture, we must consider the cognitive plausibility of the various components.  RL has always had a clear connection to cognition (e.g., \cite{schultz}), and the GA aspect (Darwinian reproducing classifiers with genetic variation) has recently been linked to concrete low-level cognitive processes through the Darwinian replication of neurological patterns \cite{replicators}.  The classifier representation is generally highly abstracted, taking the form of, e.g., real intervals \cite{Wilson1999b}, hyperellipsoids \cite{lanzi:2006:hyp}, and feedforward neural networks \cite{Bull:2002:UCN}.  In this article we use spiking neural networks \cite{spiking-n-m}, which provides our system with cognitive validity in all three LCS components.

 In this study we focus on using LCS as a cognitive architecture to solve problems that have a temporal component.  Such problems are found ubiquitously in nature, and as such it is reasonable to assume that a cognitive architecture should be able to solve them.  To do so, it must for account for the temporal nature of the environment in which it acts.  Tasks that take place in environments that contain temporal information are called semi-MDPs \cite{sutton-99}, and the ability to solve them is a precursor to complex cognitive behaviours.  Some LCS --- such as the temporal classifier system TCS \cite{journals/alife/HurstB06} --- include a mechanism to permit the solving of semi-MDPs through temporally persistent actions and a temporal RL algorithm that explicitly takes time into account when rewarding these actions.

 However, the RL is only one part of a temporal cognitive architecture, as the process of cognition itself is also temporally dynamic \cite{rhythms}.  In an LCS, this corresponds to the classifiers themselves having some temporal dynamism.  We have previously studied the use of temporally dynamic spiking neural network classifiers in the context of LCS \cite{howard2010spiking}.  Spiking networks are of particular interest as (i) they are biologically plausible as part of a cognitive architecture, and (ii) temporal dynamism may provide benefits when solving temporal problems \cite{maass}.

To date, there have been no LCS that combine temporal RL with a temporally sensitive classifier representation.  Here, we present the first LCS to couple temporally dynamic classifiers with temporal RL in an LCS framework.  The GA uses self-adaptive search parameters,  allowing it to tailor it's own learning rates in response to repeated interations with the environment.  Such flexibility is thought to be a requirement for the type of ``brain-like'' cognitive system that we wish to create.

Our hypothesis is that the coupling of temporal RL (from TCS) and neural processing (from the spiking networks) is somehow beneficial to the system.  In other words: is there a synergistic relationship between spiking networks and temporal RL that can be exploited when solving semi-MDPs?  To address this question, we test our spiking LCS on two RL benchmark problems and a robotics task.  We choose robotics-inspired tasks as they are challenging and require multiple state transtions to solve.   We show that TCS can more directly use the temporal behaviour of the neurons when solving these semi-MDPs.  Experimentation reveals that such a system is able to outperform a neural LCS that does not take neurodynamics into account.

The main contribution of this work is a hybrid combination of temporal RL with temporal neurodynamics, creating the first LCS in which the learning algorithm and classifier representation are (i) temporally sensitive, and (ii) grounded in cognitive science.

%%%%%%%%%%%%%%%%%%%%%
%%%%%%%%%%%%%%%%%%%%% BACKGROUND
%%%%%%%%%%%%%%%%%%%%%

\section{Background Research}

Learning Classifier Systems sit at the cross-disciplinary intersection between RL, ensemble learning, and evolutionary algorithms.  Our work also considers spiking networks and corresponding areas of computational neuroscience.  We focus our research on three main pertinent areas: neural learning classifier systems, spiking networks, and, as we are interested in robotic applications, evolutionary robotics.  We begin with a high-level, general description of how an LCS works.

\subsection{Learning Classifier Systems, in Brief}

Learning Classifier Systems are online evolutionary reinforcement-based learning systems that evolve a population of {\em(condition, action, prediction)} classifiers using a GA \citep{Holland75}.  Early LCS use a ternary condition, comprising binary digits \{0,1\} and a generalisation character \{\#\}, which matches certain subspaces of a binary-represented state space.  Real-valued conditions \cite{Wilson2001b} similarly provide an upper and lower bound for matching more ubiquitous continuous-valued state spaces.  Generally, the classifier representation used determines the applicability of the LCS to a given problem type.  Each classifier also has an action, which the classifier advocates in the state subspaces that it matches, and has a prediction value, which represents the payoff the classifier expects from carrying out its action in those subspaces, typically attributed through RL.

When an input state is presented to the LCS, each classifier's condition determines if the classifier is active (matches).  An action is then selected from the list of available actions advocated by the matching classifiers, which is decided by an action selection policy.  A reinforcement learning algorithm then assigns reward and updates classifier prediction values accordingly to reflect the outcome of the action.

A steady-state GA periodically generates new classifers, gradually improving the ability of the classifer population to successfully map all states and actions to payoffs.  Classifiers are selected for GA reproduction based on their fitness, which is normally based on either the prediction value, or, more recently, the accuracy of the classifiers prediction.  Accuracy-based fitness is generally a better approach, as it stops classifiers from predicting high, inaccurate values, which can damage the learning process.  A benefit of accuracy-based fitness is that the LCS eventually learns not just the optimal policy and it's predictions, but a complete and accurate {\em state, action, prediction} mapping of the entire environment.

Eventually, repeated interaction with the environment (and subsequent reinforcement learning updates and classifier improvements via the GA) produces a population that can perform optimal actions in an environment given an arbitrary input state, and generates accurate predictions for every state, action combination.

\subsection{Neural Representations in Learning Classifier Systems}

Neural networks have been used in LCS to compute predictions \cite{xcsf-neuralpred}, and as a direct replacement for classifiers.  The initial work exploring artificial neural networks as classifiers \cite{Bull:2002:UCN} used feedforward MLPs \cite{rumelhart86} --- the input state is passed to the network and used to calcualte an action.  As actions are calculated, as opposed to being static as with traditional LCS, generalisation occurs as different actions can be calculated in response to different state inputs.  Determination of matching an arbitrary state input is computed as a network output in the same way as an action.  A type of network-wide feature selection \cite{conf/gecco/HowardB08} has been shown to provide generalisation by allowing a network to selectively ignore certain inputs, which is more useful in an ensemble approach than a monolithic one as the networks tend to benefit from specialisation in the former case.

Various other network forms have since been integrated within an LCS \cite{bull-hurst-tech03,journals/alife/HurstB06}. These neural classifier systems have all been shown amenable to a neuroevolutionary approach known as constructivism \cite{QandS}, which can be defined as the gradual adding of complexity (neurons and connections) to initially-simple networks until some required level of computing power is attained.  Combined with an ensemble system --- such as an LCS --- simple, specialised structures can be quickly evolved to contribute meaningfully to solution quality \cite{nn-ensemble-rl}.  Single networks in an ensemble are typically only responsible for a part of the system's overall behaviour (e.g., operating on a reduced subset of inputs, or on a subspace of the entire problem space).  We believe this approach to be particularly suited for spiking ensembles, as monolithic spiking networks can be difficult to evolve for certain tasks.

More recently neural LCS have been used as cognitive models.  Continuous-Time Recurrent Neural Networks \cite{beer} are used as one of many possible elements in an LCS using mixed-media classifiers \cite{churchill2014evolutionary}.  Delving further into computational neuroscience, plastic spiking networks have been coupled with an LCS ``brain'', and shown to allow for a mechanistic explanation of the ``neuronal replicator'' hypothesis of Fernando \cite{replicators} (briefly, that patterns of neural activity can undergo a Darwinian evolutionary process within the brain) --- although we note that this system is only described, not implemented \cite{fernandoLCS}.

\subsection{Spiking Networks}

Spiking networks model neural activity in the brain to varying degrees of precision.  Two well-known phenomenological implementations are the Leaky Integrate and Fire (LIF) model and the Spike Response Model (SRM)~\cite{spiking-n-m}, with the most well-known mechanistic alternative being the Hodgkin-Huxley model~\cite{hodgkin-huxley}. A spiking network comprises a number of neurons connected by numerous unidirectional synapses.  Each neuron has a state, which is a measure of internal excitation, and emits a voltage spike to all forward-connected neurons if sufficiently excited.  This state is a form of memory which allows the network to solve temporal problems.  Due to spike-based communication and temporal dynamism, spiking networks are widely accepted to be the most biologically plausible neural model available. Recent research shows that even a single spiking neuron can effectively process temporal information in its input \cite{single-snn-neuron}

The evolution of spiking networks, especially for temporal problems, is a promising area of research.  Hagras and Sobh \cite{hagras2002intelligent} evolve spiking networks for online robot control.  Of particular interest is the work of Dario Floreano's group.  Full topology-plus-weight evolution is used for vision based robot control \cite{Floreano:2001:ESN}.  They also produce compact controllers for ground-based robots~\cite{evo-bns}. A common theme of this research is evidence that the inherent dynamics of spiking networks make them suited to temporal robotics tasks.

\subsection{Learning Classifier System Robotics}

The first LCS specifically for robot control appears in 1994\cite{journals/ai/DorigoC94}.  The authors create a hierarchical LCS in which lower-level LCS learn simple behaviours that high-level LCSs can then coordinate to generate complex actions.  MONALYSA \cite{Donnart1996c} allows the hierarchy itself to be dynamically reconfigured.  Experimental demonstrations show that hierarchical decomposition of behaviour into sub-behaviours produce better performance than a monolithic approach.  Fuzzy classifier systems \cite{Bonarini1998a} \cite{Pipe:2002:FRE} are shown to be capable of optimal behaviour non-trivial maze environments.

Other examples of LCS robotics include latent learning \cite{stolzmann}, which allows a robot to perform in continuous environments requiring large numbers of state transitions by using an internal environmental model to build chains of classifiers without waiting for subsequent inputs.  A simple LCS \cite{conf/cec/CazangiZF03} evolves robot controllers for goal location and object avoidance tasks in unknown environments.  Ongoing research by Butz et al. (e.g., \cite{conf/gecco/ButzH08}) demonstrates an LCS that can control a robot arm.

Perhaps the most popular algorithm for LCS robot control is the temporal classifier system --- TCS --- which has previously been implemented in LCS employing both strength-based \cite{conf/cec/StudleyB05} and accuracy-based \cite{journals/alife/HurstB06} fitness.  TCS posesses two key features that make it suitable for temporal problems.  Firstly, it has the ability to chain together multiple actions which are automatically executed on receipt of the first input state and can persist through extended periods of time.  Secondly, a temporal reinforcement learning algorithm allows the system to reward optimally-long action chains.  As TCS can search a for optimal temporal actions, the need to predefine such actions is removed; TCS can therefore automatically perform temporal state decomposition across large numbers of state transitions.

Further examples of TCS robot control include a demonstration of the ability to disambiguate between aliased states using an internal memory register \cite{conf/ecal/WebbHRL03}.  More recent research \cite{conf/iwcls/MoioliVZ07} compares results of a T-maze navigation task, again using a memory register to handle aliased states.  Results show strong performance in both real and simulated robotics tasks.  TCS has also coupled with MLP neural networks \cite{howard-gecco09}, although these are simply feedforward input-to-output mappings and as such do not possess any true neural processing ability.  To date, no TCS has employed a temporally dynamic classifier represention.

In summary, we have shown that:

\begin{itemize}
\item Spiking networks are biologically plausible processors of temporal information.
\item Spiking robotic controllers have previously been demonstrated to show high performance in semi-MDP robotics tasks.
\item LCS robotics has a strong track record of success, especially in the case of the TCS classifier system.
\end{itemize}

This section is intended to motivate experimentation on a logical extension of TCS, where for the first time a temporally-sensitive classifier representation is coupled with a temporal reinforcement learning scheme.  This approach aims to move towards a more powerful, plausible cognitive architecture based on the rich intrinsic neural dynamics of spiking classifiers.

%%%%%%%%%%%%%%%%%%%%%
%%%%%%%%%%%%%%%%%%%%% IMPLEMENTATION (NO REFS!!!!!)
%%%%%%%%%%%%%%%%%%%%%
\section{Implementation}
\label{implementation}
To clarify some important nomenclature: An {\em experiment} lasts for a number of trials.  Each {\em trial} consists of a number of {\em agent steps}.  An agent step begins with the receipt of sensory input and ends with an action being generated by the network and carried out.  To generate the action, the state input is processed 5 times by each network (5 {\em network steps}), and the resulting spike trains are used to calculate the action.

As TCS can traverse multiple states following a single sensory input, we distinguish between an {\em agent step}, which corresponds to a single action taken in an environment, and a {\em macro step} which is a chain of multiple agent steps.  Note that a step in these environments is equivalent to an action.

\subsection{Spiking Classifiers}
\label{sec-Spiking-Network}

\begin{figure}[t]
 \begin{center}
 \includegraphics[height=6cm]{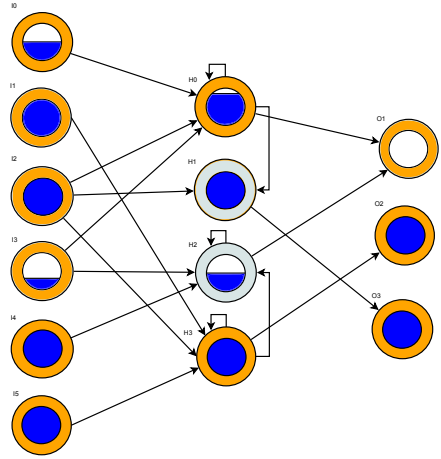}
 \caption{Showing a sample spiking network during activation.  A neuron may be excitatory (orange/dark shade) or inhibitory (grey/light shade), and has a membrane potential (blue/dark filling of the centre circle).  Membrane potential may build through a number of timesteps, eventually surpassing a threshold and emitting a spike (filled inner circle), before resetting (empty inner circle).  Each network has a problem-dependent number of inputs, initially one hidden layer neuron, and three outputs, the latter of which is the ``dont-match'' neuron.}

 \label{fig1}
 \end{center}
 \end{figure}

Discrete-time Leaky Integrate and Fire (LIF)~\citep{spiking-n-m} networks are used as spiking classifiers, fulfilling the role of condition and (calculated) action.  Three layers of neurons are connected by numerous weighted connections, which can be recurrent within the hidden layer only.  Each network has one input neuron per state input, and three output neurons.  A sample network is shown in Fig.~\ref{fig1}.

On network creation, the hidden layer is populated with a single neuron, whose type is intitially excitatory (transmit voltages $V$$\geq$ 0), otherwise it is inhibitory ($V$$<$0) (50\% chance of each).  Input and output layer neurons are always excitatory.  Each possible connection site is initially occupied by a connection, and all connections are weighted uniform-randomly in the range [0-1].

When presented with an input state, each input is scaled to the range [0-1] and applied to its respective input neuron.  Stimulation by incoming voltage (from either the input state or spikes generated by other neurons) alters a given neurons membrane potential $m$, $m>0$, which by default decreases over time.  Surpassing a threshold $m_\theta$ causes a spike to be transmitted to all forward-connected neurons, which is weighted according to the weight of the connection that the spike travels along.  $m(t)$ is the membrane potential at processing step $t$ (calculated following equation~\ref{eq3}), $I$ is the input voltage, $a$ is an excitation constant and $b$ is a leak constant.  Immediately after spiking, the neuron resets its membrane potential to $c$ following equation~\ref{eq4}.  Parameters are $a=0.3$, $b=0.05$, $c=0.0$, $m_\theta = 1.0$.

\begin{equation}
m(t+1) = m(t) + (I+a-bm(t))
\label{eq3}
\end{equation}
\begin{equation}
\mathrm{If}\; (m(t) > m_\theta)\;\;\;\; m(t) = c
\label{eq4}
\end{equation}

All tasks in this paper require discrete-valued outputs.  To generate an action, the current input state is run for 5 network steps, which generates between 0-5 spikes at each of the three output neurons.  We count the number of times each output neuron spikes, and classify that neurons activity as either {\em high} (3 or more spikes generated out of 5) or {\em low} (less than 3 spikes generated).  Each task maps these activations into actions in a different way.  The final output neuron is a ``don't match'' neuron that excludes the classifier from the match set if it has high activation.  This is necessary as the action of the classifier must be re-calculated for each state the classifier encounters, i.e. a classifier may match differently/advocate different actions in different state subspaces.

\subsection{Spiking TCS}
\label{sec-The-System}

Spiking TCS (STCS) is derived from the accuracy-based TCS \cite{journals/alife/HurstB06}, with the additions of computed prediction \cite{Wilson2001a} and spiking classifiers.  STCS comprises a population of classifiers, which in this case are spiking networks.  Each classifier has a prediction weight vector which is used to compute its prediction value.  Note the distinction between ``connection weight'', which refers to a neural network weight, and ``prediction weight'', which is used  to calculate classifier prediction. Each classifier has various other parameters, including the last time it was involved in GA activity, the accuracy of its prediction value, and the fitness of the classifier.

STCS solves RL problems by controlling the actions of an agent in a task-dependent state space until either a reward is received, or a time limit is reached.  STCS uses two types of trial, explore and exploit, which are carried out alternately until some predefined trial limit is reached and the experiment ends.  Exploration allows STCS to evaluate the benefits of carrying out various state, action combinations, and to incrementally build a payoff mapping of the environment.  Exploitation tests the current ability of STCS to solve the problem.

A trial commences with the reciept of an initial state $s_t$, which is processed by each network to calculate matching and advocated action.  All classifiers that match the state are put into a match set [M].  Full exploration of the state space, which is key to the learning ability of the system, can be hindered if some actions are missing from [M].   If this is the case, {\em covering} repeatedly generates random networks until the newly-created network action equals the missing action.

Classifier prediction $cl.p$ is then computed as a linear combination of the prediction weight vector $w$ and the current state $s_t$, which promotes generalisation by allowing the same classifier to accurately predict different values in different states (equation~\ref{pred1}).  The weight vector is always one element larger than the state vector, with all elements initially 0.  Note that the first element of $w$ is multipled by a constant $x_0$.

\begin {equation}
 cl.p(s_t) = cl.w_0 * x_0 + \sum_{i>0} cl.w_i * s_{t(i)} \label{pred1}
\end {equation}

Classifier predictions are weighted by their fitness, and averaged for each action to provide a mean estimated payoff per action (creating the prediction array).  STCS then selects a {\em system action} from the prediction array.  This action is selected payoff-proportionately in explore trials \cite{journals/alife/HurstB06}, to provide some exploration without encouraging time-wasting random action selection (which is the traditional policy \cite{Wilson2001a}).  In exploit trials, the highest-payoff system action is deterministically selected.  All classifiers that advocate the selected system action are put into the action set [A].  The system action is then carried out.

\subsubsection{The Drop-decision Cycle}

As with other TCS, STCS allows [A] to control the agent for more than one agent step by forming extended macro steps.  Once the first action is taken, STCS enters the {\em drop-decision} cycle.  The next input state is fed into [A], and the spiking classifiers in [A] recalculate their actions/matching.  If all classifiers in [A] still advocate the system action given this new input, [A] persists in controlling the agent.  If no classifiers match the state, or a timeout $t_{drop}$ is reached, the drop-decision cycle is terminated.  If only some classifiers still match, proportional fitness-weighted payoff selection picks a classifier from [A]:

 \begin {itemize} 
\item If the classifier matches we continue with [A], removing non-matching classifiers.
\item If the classifier does not match, we remove matching classifiers from [A], perform necessary parameter updates, then exit the drop-decision cycle.
\end {itemize}

If classifiers in [A] advocate different actions, a ``winning'' system action is picked from [A] in the same way.  All classifiers that do not advocate the newly-selected system action are removed from [A].  Classifiers removed in this way are not candidates for parameter updates; as the outcome of using their action is not explored, an accurate prediction value cannot be ascertained.

\begin{figure}[t!]
\begin{center}

 \psfig{file=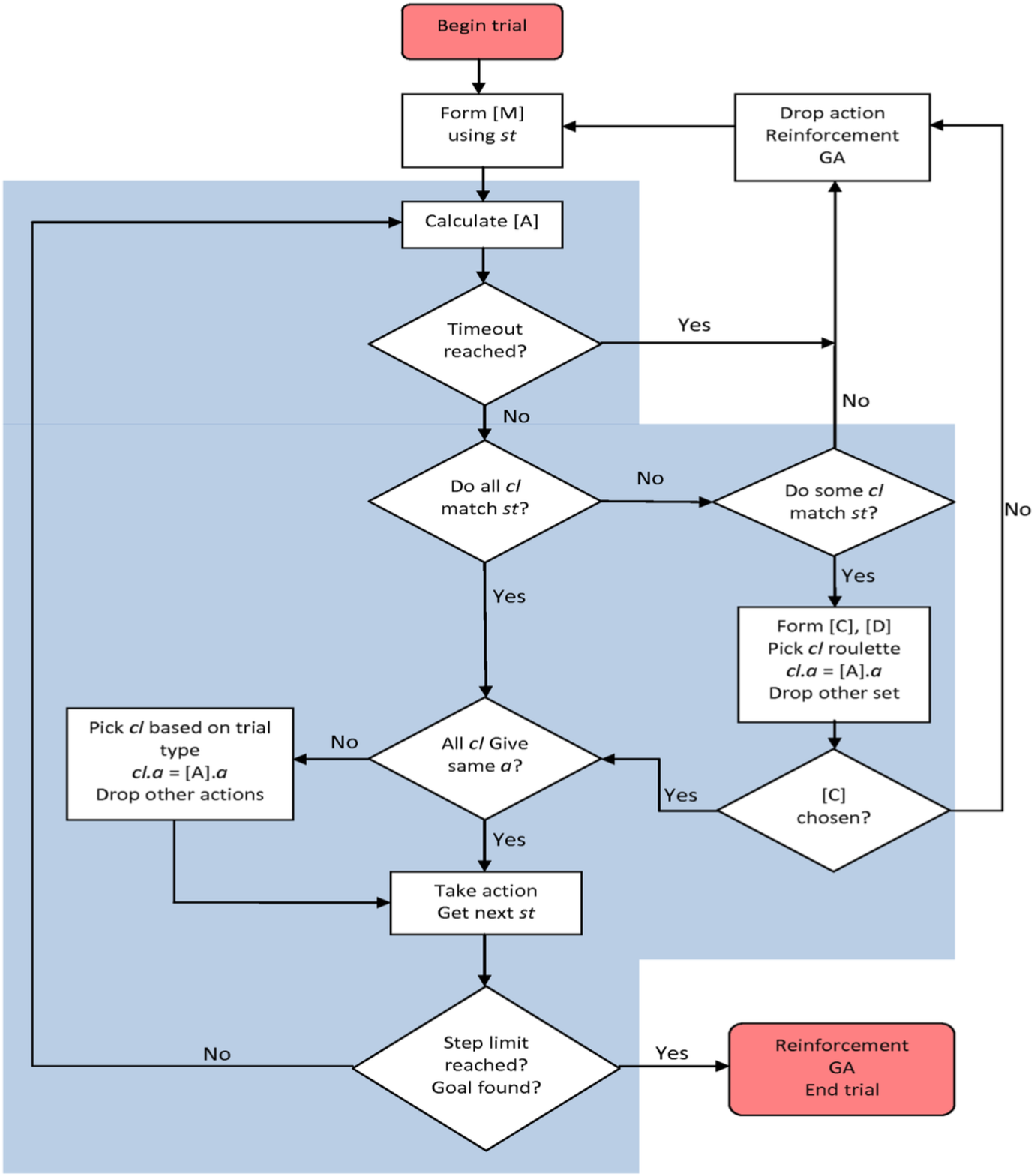,width=10cm,height=15cm}

\end{center}
\caption[]{Flowchart showing the TCS drop-decision cycle.}
\label{tcs-flowchart}
\end{figure}

Following this, the current [A] is saved as [A$_{-1}$].  A new [M] is formed and STCS proceeds to calculate predictions, select a system action, and re-enter the drop-decision cycle.  The TCS algorithm is shown in Fig.~\ref{tcs-flowchart}.  This process continues until reward is returned or a timeout is reached, which ends the trial.

When the drop-decision cycle terminates, external reward is potentially returned from the environment if the goal state is reached, and the classifiers remaining in [A] are updated.  Classifiers in the previous action set [A$_{-1}$] may receive a discounted reward.

\subsubsection{Temporal RL Component}
%REINFORCEMENT
Traditionally, reinforcement in LCS is based on Q-learning \cite{watkins-thesis}.  This is unsuitable for semi-MDPs as it does not take time into account when attributing reward.  Equation \ref{eq2} shows the TCS reinforcement forumula.  $P$ is the target prediction, $r$ is external reward, and $maxP$ is the highest prediction array value for the current state.

The first reward factor, $e^{- \varphi t^t}$, discounts the immediate external reward by a function of $t^t$, the total number of agent steps taken in the trial.  The second reward factor, $e^{- \rho t^i}$, discounts any delayed reward by a function of $t^i$, the number of agent steps taken since the last [M] formation.  The reinforcement therefore acts to reward efficient overall solutions, and encourage a smaller number of long macro steps per trial.  $\varphi$  and $\rho$ are experimentally-determined discounting factors.

\begin {equation}
P = (e^{- \varphi t^t}) r + (e^{-\rho t^i}) \times maxP
\label{eq2}
\end {equation}

Next, a classifier's prediction value is updated by altering the prediction weight vector $w$ of each classifier in [A] using a version of the delta rule (equations\ref{delta_wi} and \ref{alter_wi}).  Note that P is the target prediction value as calculated from the temporal RL formula above.  Prediction error is then updated (equation \ref{error}) --- $\epsilon$ is the classifier's error and $\beta$ is a learning rate.

\begin {equation}
\label{delta_wi}
\Delta cl.w_i = \frac{\eta}{||x_{t-1}||^2}(P - cl.p(s_{t-1})) x_{t-1}(i) \\
\end {equation}

\begin {equation}
\label{alter_wi}
cl.w_i \leftarrow cl.w_i + \Delta w_i\\
\end {equation}

\begin {equation}
\label{error}
\epsilon  \leftarrow \epsilon+ \beta (|P - cl.p(s_{t-1})| - \epsilon)\\
\end {equation}

STCS uses macroclassifiers to enhance computational efficiency.  A macroclassifier represents a number of identical classifiers, and its numerosity parameter indicates the number of identical classifiers that the macroclassifier represents.  Subsumption is the mechanism by which macroclassifiers are formed --- one classifier is subsumed by another if its is less general than the subsumer and advocates the same action.  If a classifier is subsumed, it is deleted from the population and the subsumers numerosity increases by 1.  This check for generality is easy for simple classifier representations such as intervals and ternary alphabets, as it is obvious how general such a classifier is.  For spiking classifiers the process becomes computationally taxing as an exhaustive search must be carried out, computing all classifier outputs for a given state space.

There are normally two forms of subsumption \cite{Wilson2001a}, action set subsumption and GA subsumption.  We use GA subsumption, which is limited by having two parents and two children = 4 exhaustive comparisons per GA cycle.  We action set subsumption as it requires checking a classifier in [A] against all other members of [A], which is computationally infeasible as [A] can contain hundreds of classifiers.

Each classifer has a record of the last time it was involved in a GA, $ts$.  Once external reward (or a drop decision) has been signalled, the GA will  activate in [A] (or if no external reward is signalled, [A${-1}$]) if the average $ts$ in [A] ([A${-1}$]) exceeds some threshold $\theta_{GA}$.

\subsection{Neuro-evolutionary Algorithm}

STCS uses a steady-state niche GA, which activates within an action set.  Two parents are selected and used to create two offspring.  The offspring are inserted into the population, and two classifiers are deleted if the maximum population size is reached.  Classifiers are only created in [A], but deletion occurs from the popuation as a whole, therefore there is an implicit pressure towards generality as the best chance of survival is to appear in as many [A] as possble.

Each classifier has it's own self-adaptive mutation rates, which are initially seeded uniform-randomly in the range [0,1] and mutated as with an Evolution Strategy~\citep{rechenberg} as they are passed from parent to child following equation~\ref{eq-sa}.  

\begin{equation}
\label{eq-sa}
\mu \rightarrow \mu\; exp^{N(0,1)}
\end{equation}

This approach is adopted as it is envisaged that efficient search of weights and neurons will require different rates, e.g., adding a neuron is likely to impact a network more than changing a connection weight, so less neuron addition events than connection weight change events are likely to be desirable.  Self-adaptation is particularly relevant when considering the system as a cognitive architecture --- brainlike systems must be able to autonomously adapt to a changing environment and adjust their learning rates accordingly.

The genome of each network comprises a variable-length vector of connections and a variable-length vector of neurons.  Different classifier parameters govern the mutation rates of connection weights ($\mu$), connection addition/removal ($\tau$), and neuron addition/removal ($\psi$ and $\omega$).  For each comparison to one of these rates a uniform-random number is generated; if it is lower than the rate, the variable is said to be {\em satisfied} at that allelle.  During GA application, for each connection, satisfaction of $\mu$ alters the weight by $\pm$0-0.1.  Each possible connection site in the network is traversed and, on satisfaction of $\tau$, either a new connection is added if the site is vacant, or the pre-existing connection at that site is removed.  $\psi$ is checked once, and adds or removes a neuron from the hidden layer based on satisfaction of $\omega$.  New neurons are randomly assigned a type (inhibitory/excitatory, 50\% chance of either), and each connection site on a new neuron has a 50\% chance of having a connection. New connections are randomly weighted between 0 and 1.

For clarity, we now summarise the steps involved in a GA cycle.  First, two offspring networks are selected.  The parameters for those networks are self-adapted.  Connection weights are altered based on $\mu$, then node addition/removal takes place based on $\psi$ and $\omega$.  Finally, connections are added or removed based on $\tau$.  These networks are inserted into the population and networks are deleted based on a combination of their fitness and the sizes of the action sets they participate in.  This encourages the population to spread approximately evenly around the state space.

%%%%%%%%%%%%%%%%%%%%%
%%%%%%%%%%%%%%%%%%%%% EXPERIMENTATION
%%%%%%%%%%%%%%%%%%%%%

\section{Experimentation}

We selected three continuous-state semi-MDP test problems to act as benchmarks with which to assess various important qualities of the system.  Ten experimental repeats per test problem are used to create averages.  The current state of the system is saved every 50 trials and used to perform statistical analysis of results.  When comparing the STCS to other systems, twin-tailed t-tests are used to assess statistical significance (significance indicated at P$<$0.05).

Neuron membrane potentials $m$ are only reset once once at the beginning of each trial.  Preserving neuron states in this manner allows us to exploit the underlying temporal information in these semi-MDPs, which we believe is a key determinant of the performance of the system.

Each experiment is comprised of a problem-dependent number of trials, which alternate between explore and exploit.  A trial starts with the agent randomly positioned in that environments start zone, consists of a number of [M] and [A] formations as the agent navigates the environment, and terminates with either the receipt of external reward when the agent reaches the goal state or a timeout after 200 macro steps.  For all experiments, averages are taken from the entire population for all experimental repeats.

In the first experiment, we assess the ability of STCS to form appropriately-sized action chains.  The second experiment demonstrates the scalability of the system to very fine-grained state spaces.  Experiments 1 and 2 are also standard RL test problems.  The final experiment shows the ability of the system to successfully solve a noisy robotics task.  The overarching goal of the three experiments is to demonstrate the effectiveness of combining neural information processing (spiking classifiers) with temporal RL (TCS reinforcement) on a series of semi-MDPs.

Spiking N-XCSF is parameterized as follows:  learning rate $\beta$=0.2 $ (0<\beta \leq1)$,  GA threshold $\theta_{GA}$=50, deletion threshold $\theta_{DEL}$=50, XCSF constant $x_0$=1, XCSF learning rate $\eta$=0.2 $(0<\eta \leq1)$.  All other XCSF parameters follow \cite{Wilson2001a}. TCS parameters are $\varphi$=0.45, $\rho$=0.005 --- $t_{drop}$ varies between experiments. All networks initially have a single hidden layer neuron.

\subsection{Experiment 1: Mountain-Car}

This test problem is selected to assess the ability of the spiking classifier system to correctly switch between different actions in a single [M] formation.  Dropping and reforming [M] is discouraged --- each drop takes up an extra step, and we ideally want the classifiers to recalculate their action within the current [A].

The mountain-car problem \cite{sutton-scc} is a classic RL problem, in which a car must be guided out of a one-dimensional valley.  In many cases, the car must move away from the goal state to attain enough momentum to climb out of the valley, meaning that it must switch actions.  State variables are {\em position} [-1.2, 0.6], and {\em velocity} [-0.07, 0.07].  Three actions are available: forward (increase velocity), backward (decrease velocity), and no movement (maintain velocity).  As three actions are possible, output neuron activations are mapped onto actions as follows: forward = {\em high, high}, backward = {\em low, low}, and no movement = {\em high, low} or {\em low, high}.

Each experiment consists of 5000 trials, 2500 explore and 2500 exploit, with population size $N$=1000 and $t_{drop}$=20.  The agent is initally randomly placed in the environment (except in the goal state) with a random velocity, and has to reach the goal state (where the cars position is $>0.5$) in the fewest possible macro steps (average optimal=1.6).

\begin{figure*}[t!]
\begin{center}

\subfloat[]{ \psfig{file=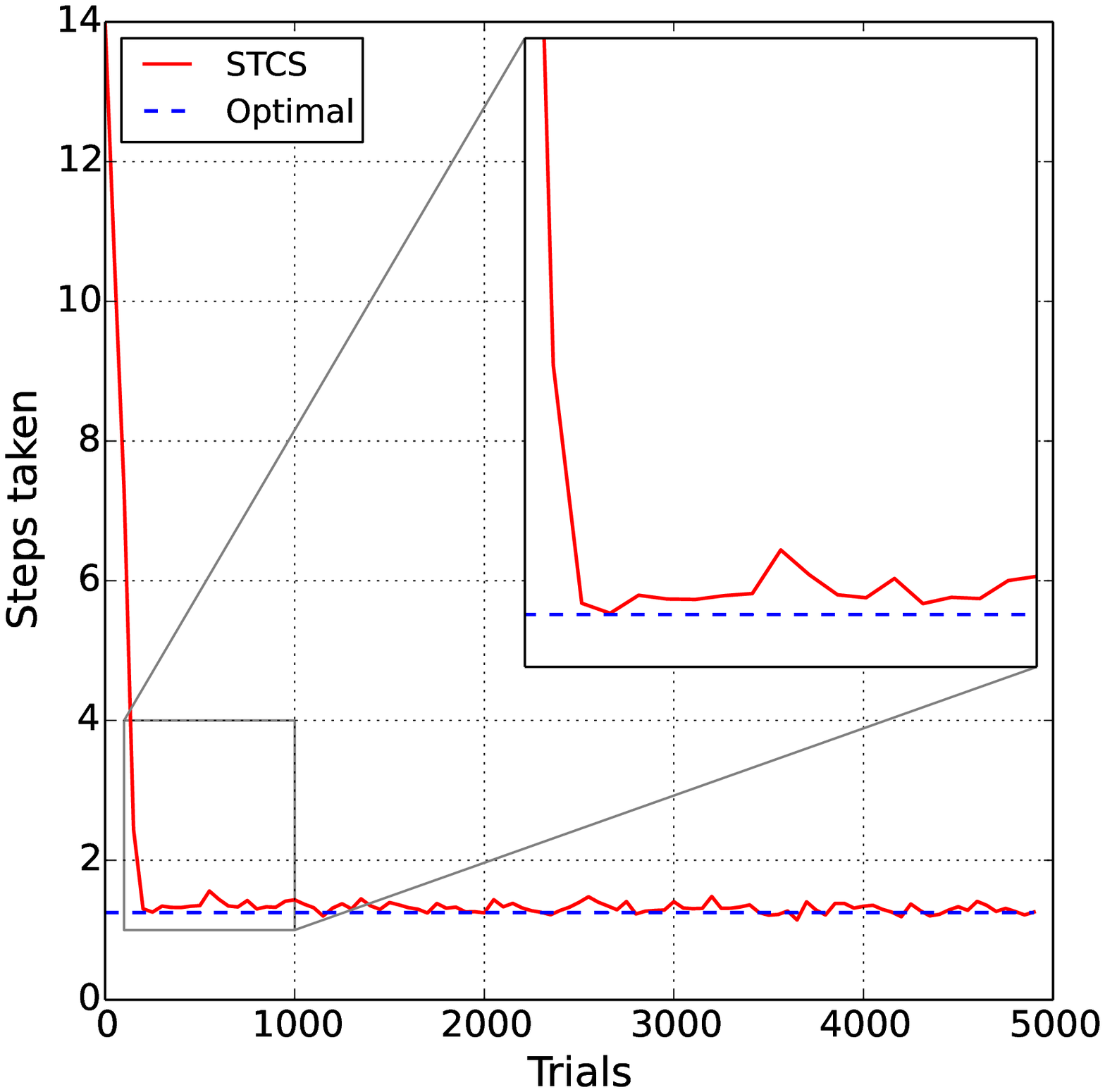,width=6cm,height=4cm}}
\subfloat[]{ \psfig{file=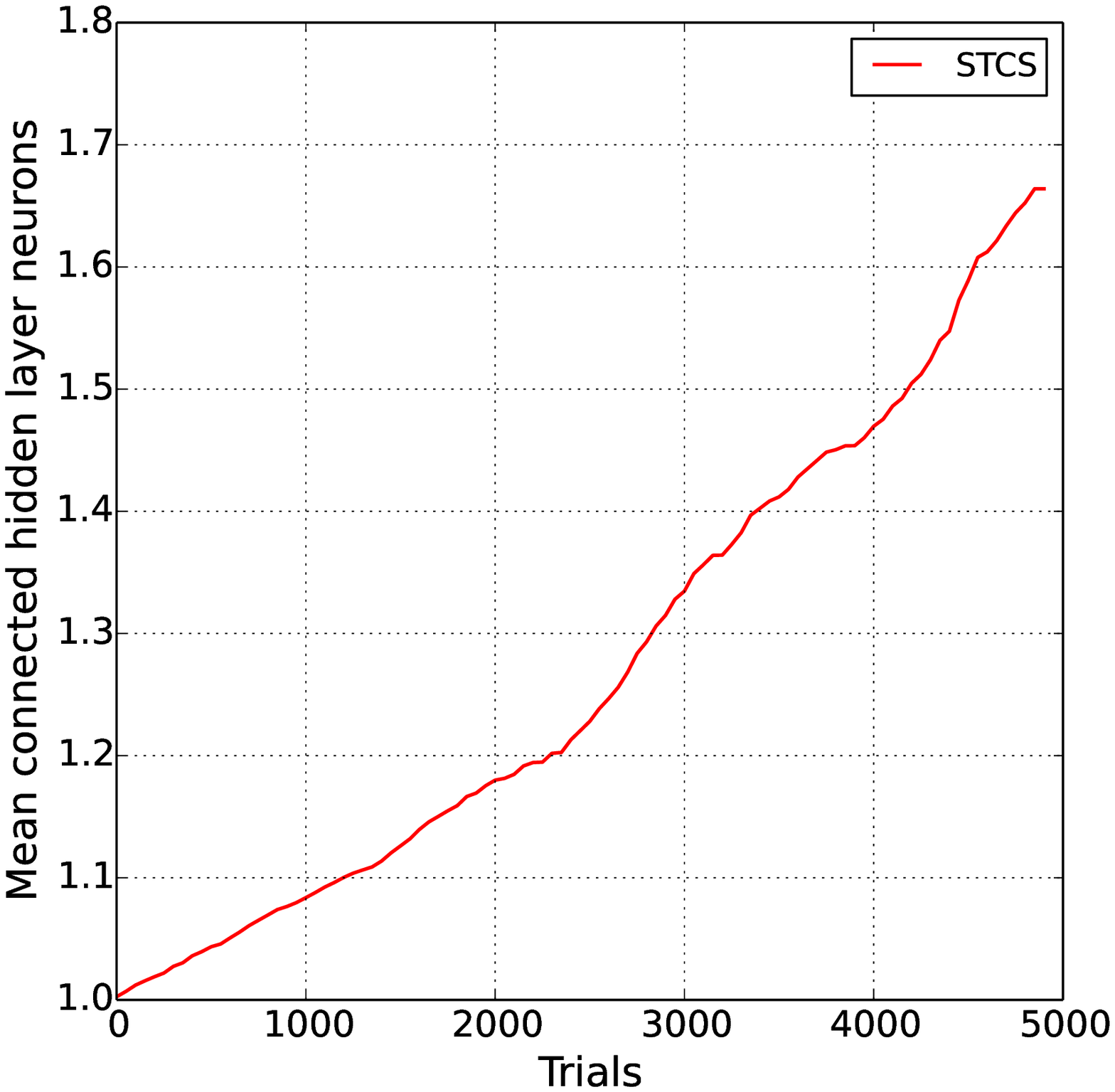,width=6cm,height=4cm}}\\
\subfloat[]{ \psfig{file=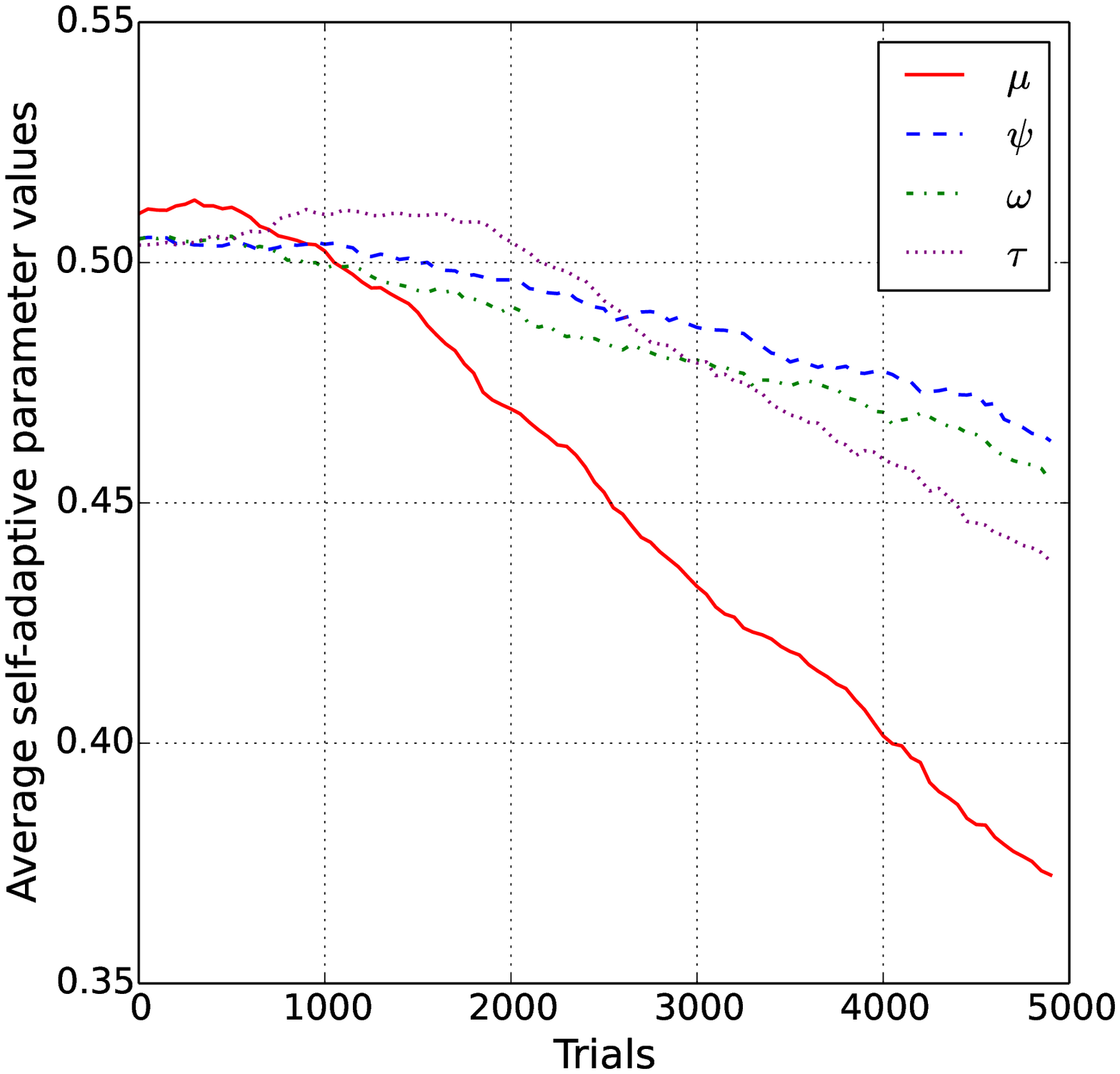,width=6cm,height=4cm}}
\subfloat[]{ \psfig{file=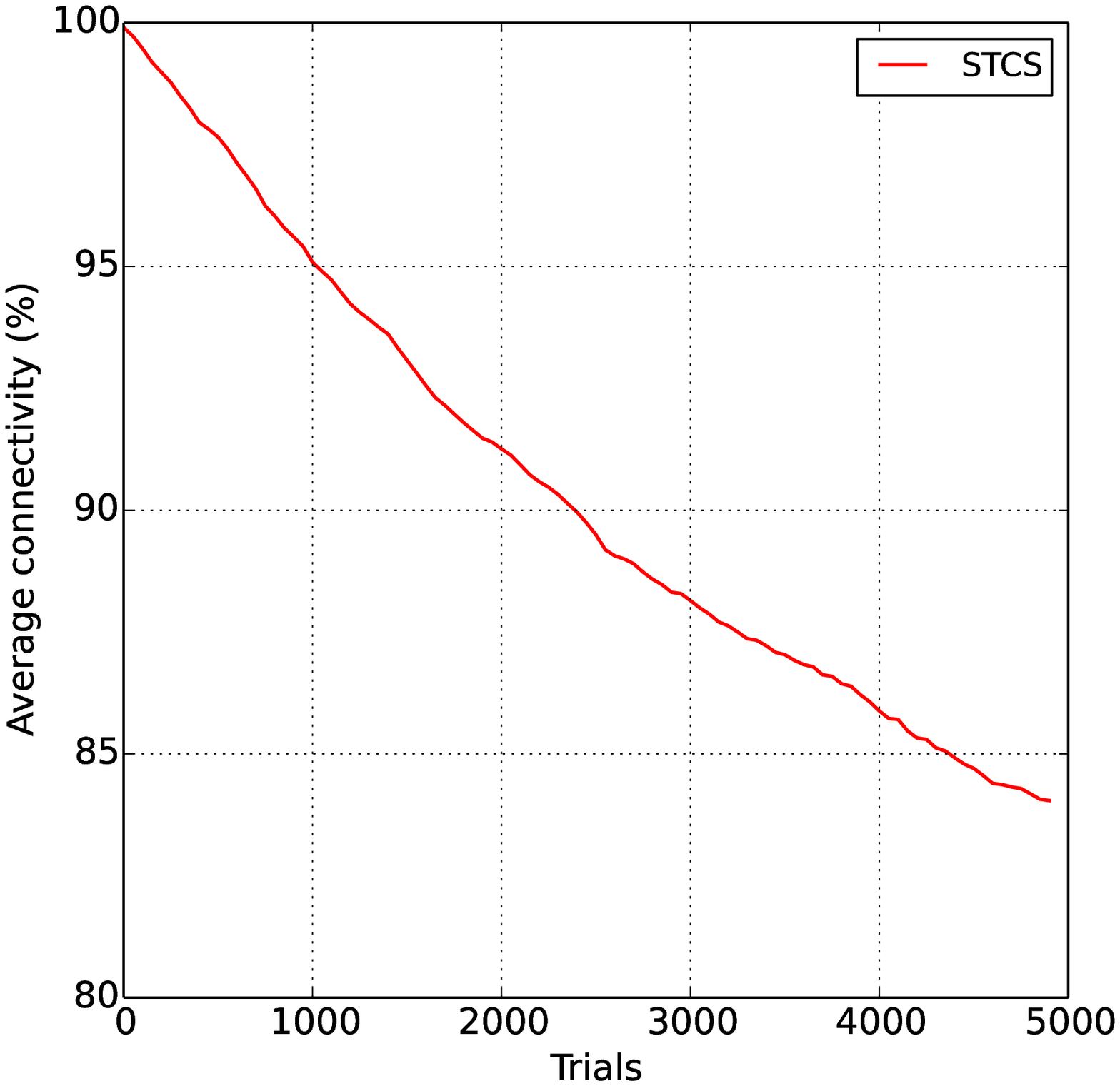,width=6cm,height=4cm}}

\end{center}
\caption[]{Mountain-car (a) average macro steps, (b) average connected hidden layer neurons, (c) average self-adaptive parameter values, (d) average enabled connections in STCS}
\label{mcar}
\end{figure*}

\subsubsection{Results}

Fig.~\ref{mcar}(a) shows attainment of optimum state space segregations within 100 trials.   The TCS reinforcement formula allows for the generation of shortest-length atomic action chains within 480 trials, indicating more expedient convergence than XCSF with tile coding \cite{lanzi-tile} ($\approx$2000 trials to convergence) with the same population size.  This is due to the ability of STCS to quickly explore the space of {\em (state, action) } combinations by chaining together actions during exploration trials.

Average neurons per classifier (Fig.~\ref{mcar}(b) gradually increases to a final value of ~1.65.  Self-adaptive parameters decline from their initial values (Fig.~\ref{mcar}(c)).  Final network connectivity (Fig.~\ref{mcar}(d)) steadily declines to a final average value of 84.5\%.  Action set analysis shows that STCS allows the agent to reach the goal state from any intial position/velocity combination by forming the minimal number of match sets.  Actions are altered from forward to backward as the agent builds momentum; this is achieved by the networks recalculating their actions through a combination of the current state input and the network's own internal state to alter the chosen action.  

\subsection{Experiment 2: Continuous Grid World}

The second experiment takes place in the continuous grid world, \cite{boyan.moore-1995:gener}, a benchmark RL test problem.  This environment is more complex than the mountain-car problem (larger state space, more actions to choose from), although it does not necessarily require the amount of action switching seen in that problem.  It is chosen as a step towards real robotics problems, whilst simultaneously allowing us to compare to previous work using a MLP neural networks in an identical classifier system \cite{howard-gecco09}.  The main difference between the two systems is that the spiking networks have a temporally-sensitive internal state, whereas the MLP networks do not.  The purpose of this experiment is therefore to ascertain the benefits of coupling a richer neural representation (capable of temporal information processing) with the temporal RL component of TCS.  We also  compare to a modern temporally dynamic classifier system \cite{DAFDGP}.

MLP networks are handled in much the same way as the spiking networks, except they process each input state once in a feedforward manner, and each neuron has an associated bias (bounded [0-1]), which is mutated as a connection weight.  Connection weights are bounded [-1,1]. As MLPs generate continuous-valued outputs, output neuron activations are {\em low} with output $<$0.5, and {\em high} otherwise.  See \cite{howard-gecco09} for full implementation details. MLP networks are a competitive benchmark, with a track record of high performance on reinforcement learning and problems \cite{he2007reinforcement}.

The grid world is an enclosed 2D arena with $x,y$ dimensions bounded [0-1].  An agent is intiallty placed anywhere except the goal, and must navigate to the goal $(x+y >1.90)$ in the fewest possible macro steps (average optimal 1.5), whereupon it receives an immediate reward of 1000 and the next trial begins.  All other outcomes give an immediate reward of 0. The state $s_t$ is defined as the agents $(x,y)$ position.  To more closely emulate the noise inherent in robotics tasks, the $x$ and $y$ positions of the agent are subject to noise; +/- [0\%-5\%] of the true position.  The activations of the first two output neurons translate to a discrete movement of length 0.05 as follows: (high, high) = North, (high, low) = East, (low, high) = South, and (low, low) = West.

As this is a comparative experiment, we measure the {\em stability} of a solution.  After each exploit trial, we perform an additional exploit trial from a static location (0.25, 0.25) and record how many steps it takes to reach the goal state.  We say that stability has been attained if the optimal number of steps is achieved in 50 consecutive static-location trials.  Stability indicates how quickly the LCS solves the problem.

Each experiment lasts for 20000 trials and uses a population $N$=20000.  All other parameters are identical to those used in the mountain-car problem.

\subsubsection{Comparing to Other LCS}

We initially compare to XCSF with real-interval classifiers \cite{conf/cec/LanziLWG05a}, which is shown to converge within $\approx$15000 trials with a population size $N$=10000.  We note the formation of optimally-long action chains in STCS within 5000 trials, with an optimal number of macro-actions being used within 2000 trials --- STCS is therefore at least competitive with this system despite the more complex classifiers it employs.  XCSF with tile coding has been shown to converge more expediently than STCS \cite{lanzi-tile} ($N$=20000), although results are critically dependent on the resolution of the coding and are worse than STCS in the worst case, varying between 500 and $>$5000 trials to convergence.

 We additionally compare to a modern LCS based on XCSF that uses fuzzy Dynamic Genetic Programming (DGP) classifiers \cite{DAFDGP} with the same population size $N$=20000.  This has been chosen as, much like the spiking classifiers used in STCS, DGP classifiers allow for temporal dynamism.  Results (Fig.9 in \cite{DAFDGP}) show the attainment of optimum performance within $\approx$30000 trials.  STCS converges more quickly than the comparative classifier systems thanks to the TCS mechanism, which allows for instant generalisation over the search space and less constrained environmental exploration during exploration trials.

\subsubsection{Comparing STCS to MLP TCS}

\begin{figure*}[t!]
\begin{center}

\subfloat[]{ \psfig{file=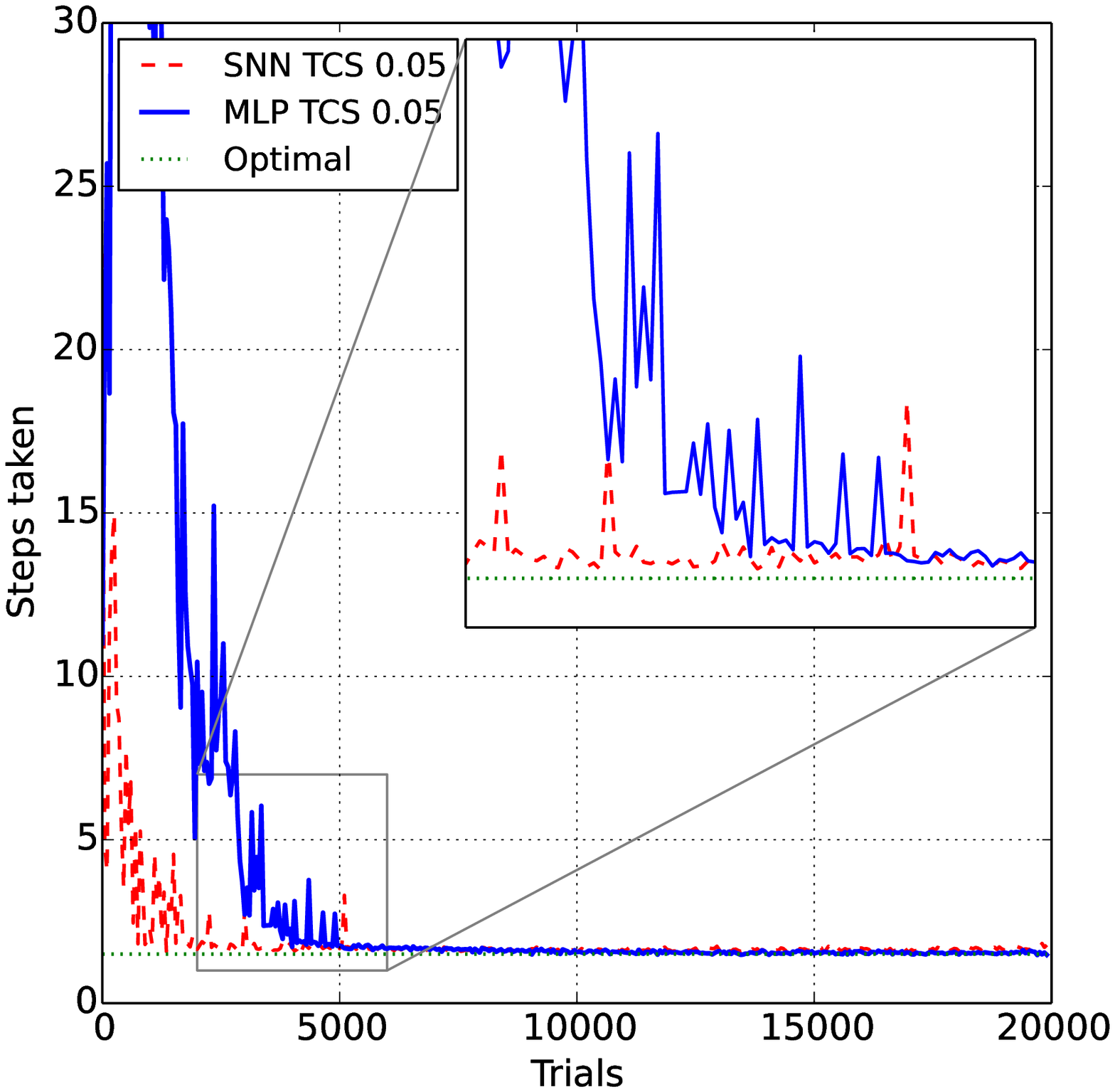,width=10cm,height=5cm}}\\
\subfloat[]{ \psfig{file=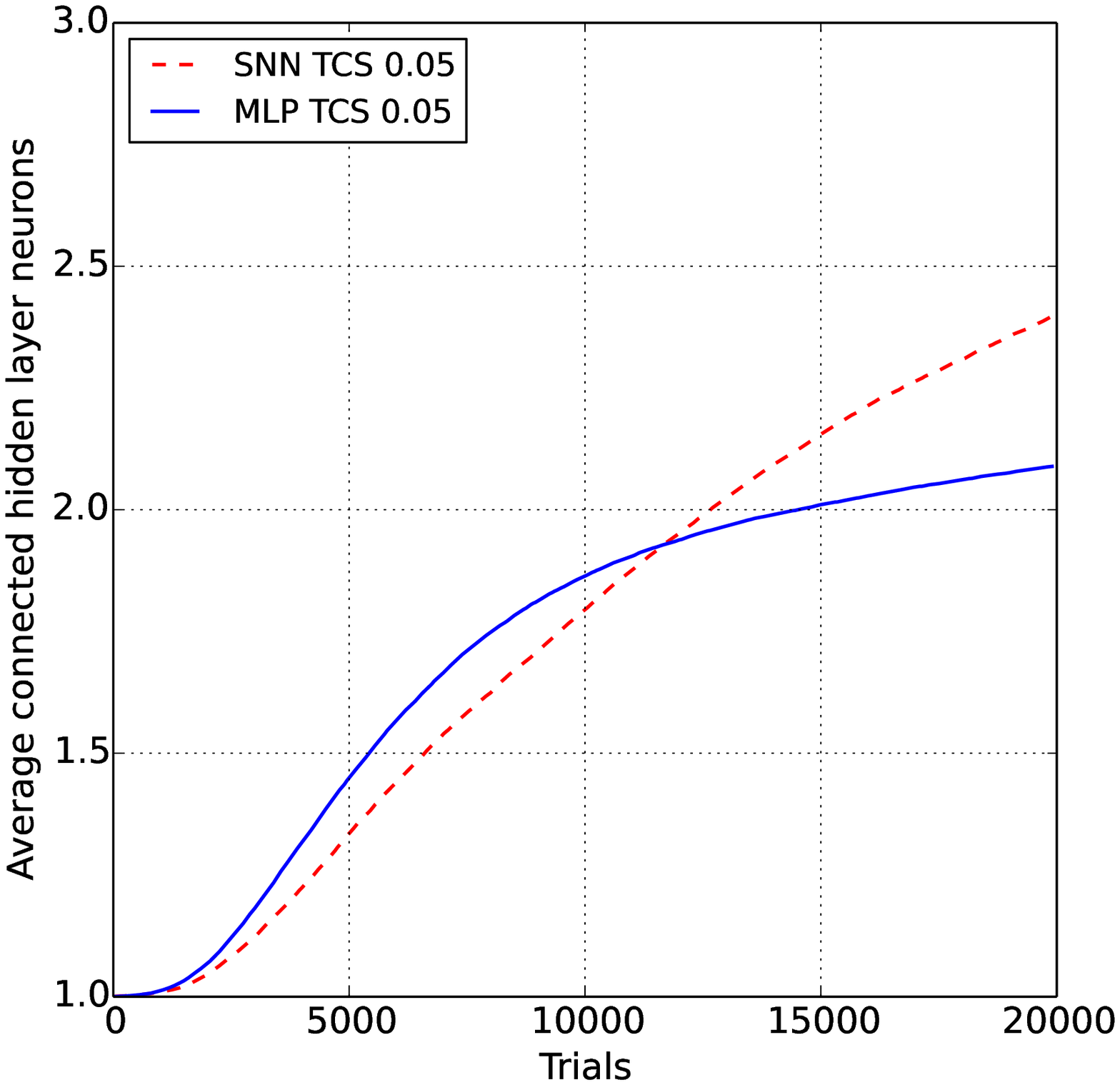,width=6cm,height=4cm}}
\subfloat[]{ \psfig{file=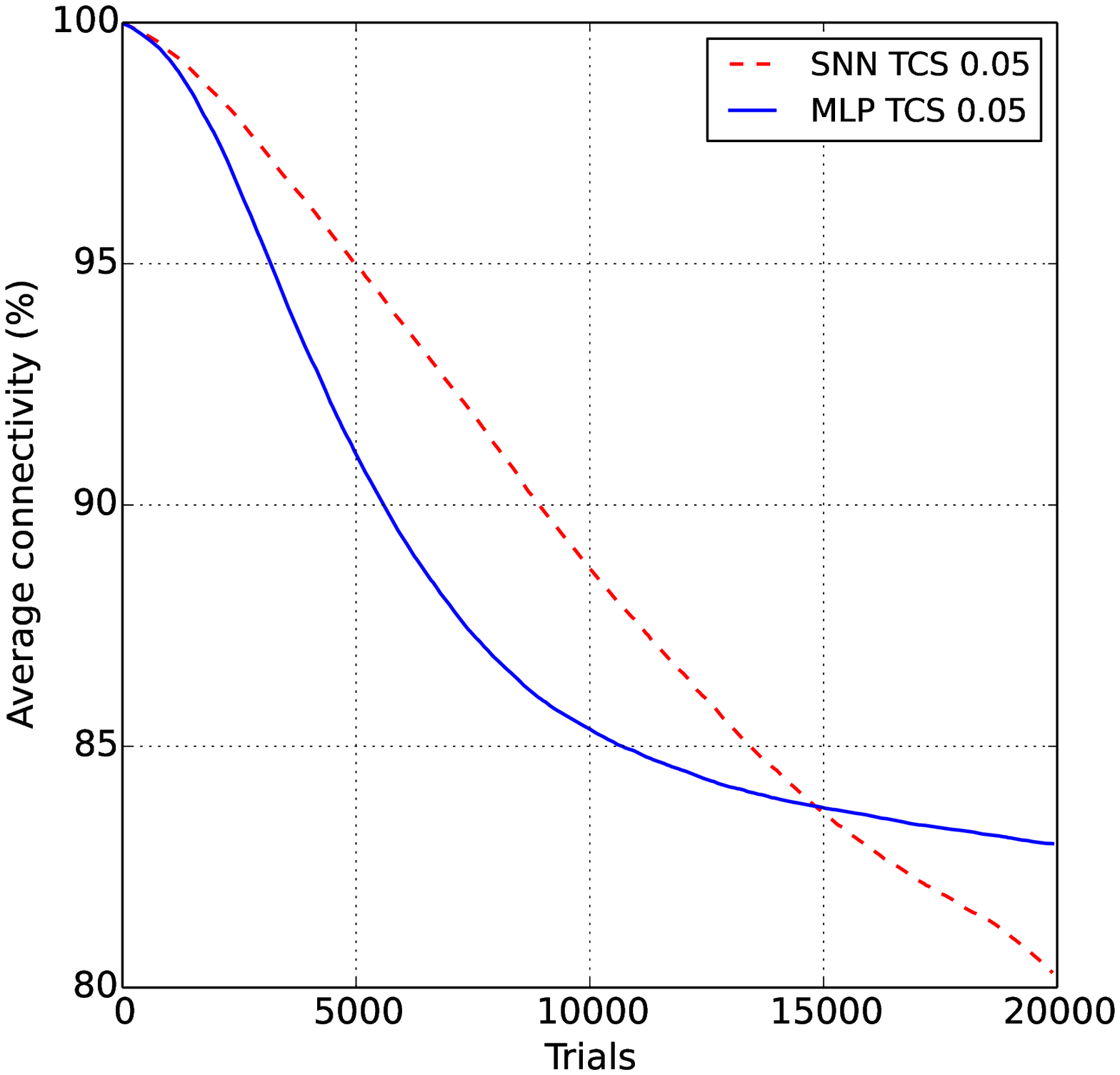,width=6cm,height=4cm}}\\
\subfloat[]{ \psfig{file=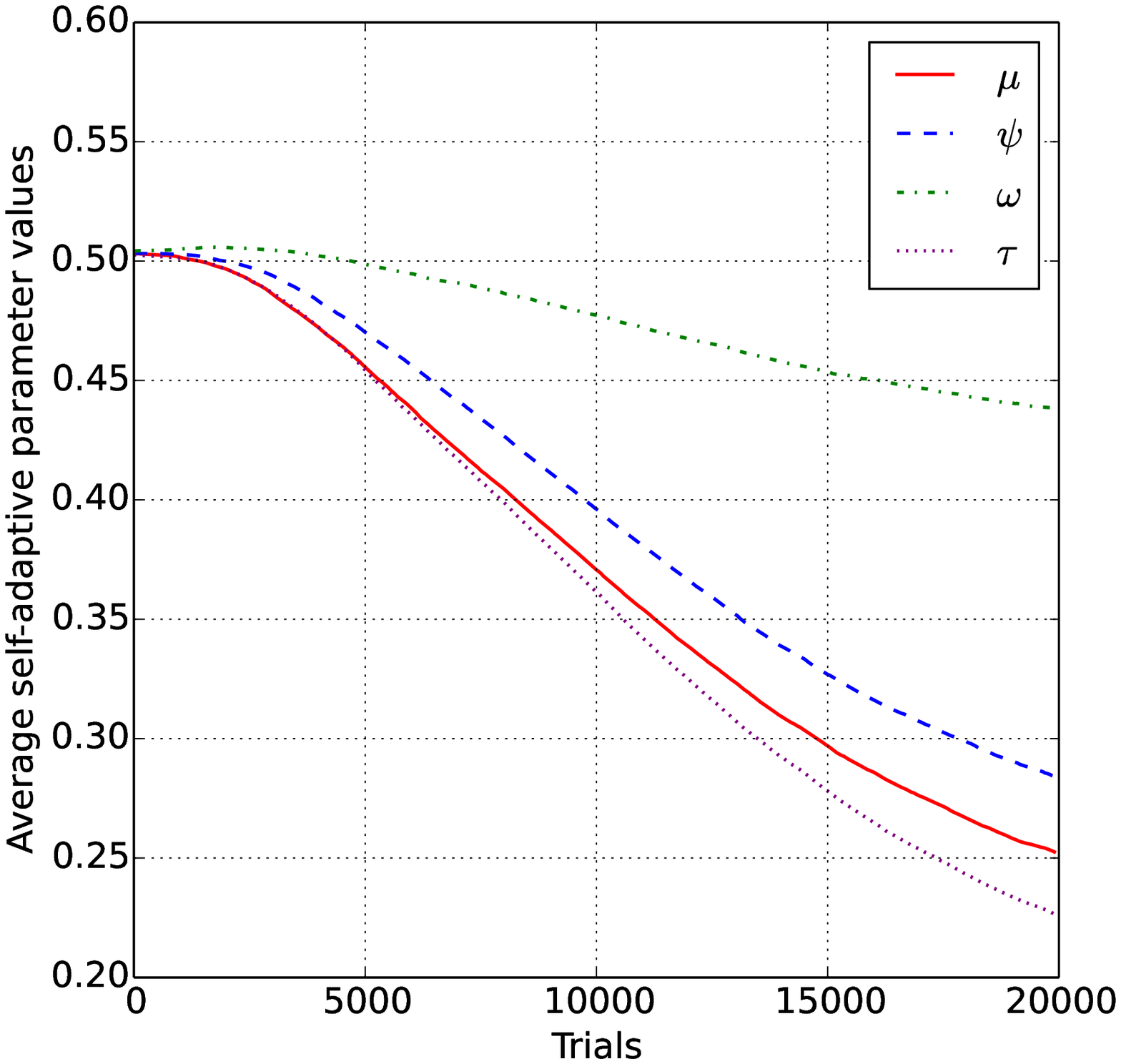,width=6cm,height=4cm}}
\subfloat[]{ \psfig{file=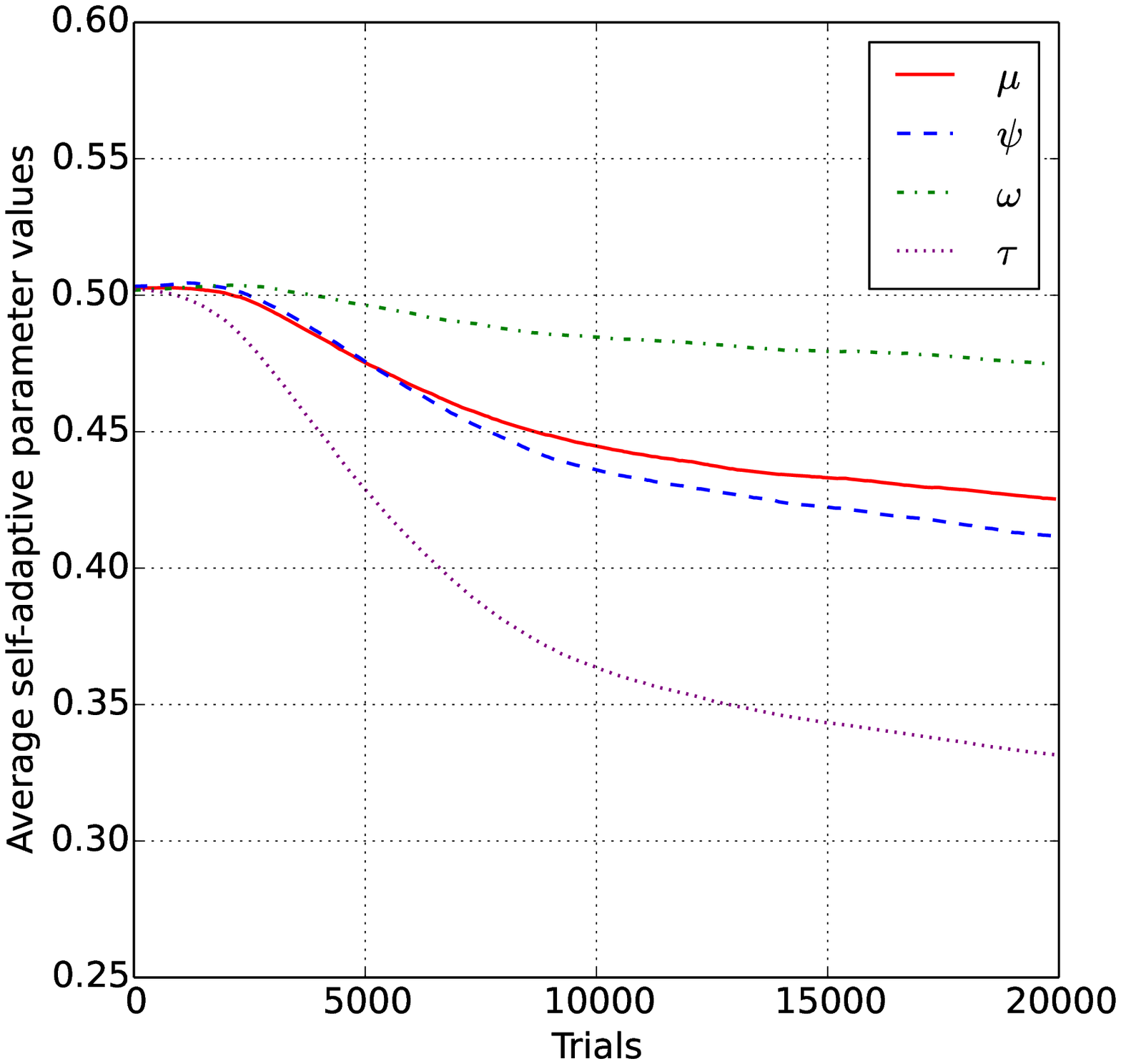,width=6cm,height=4cm}}

\end{center}
\caption[]{Continuous grid world 0.05 (a) average macro steps, (b) average connected hidden layer nodes, (c) average enabled connections, (d) average STCS self-adaptive parameter values, (e) average MLP TCS self-adaptive parameter values.}
\label{Spike-TCS}
\end{figure*}

Fig.~\ref{Spike-TCS}(a) shows optimal performance after 2000 trials for STCS, compared to 5000 trials in the MLP case.  The spiking classifiers contain more hidden layer neurons on average (2.4 vs 2.09), and are slightly less connected (81\% vs 84\%) (Fig.~\ref{Spike-TCS}(b)/(c)), however neither of these differences are statistically significant.

The spiking representation holds one main advantage over the MLP representation in this environment, having a statistically lower number of macroclassifiers (16552 vs 18148, P$<$0.05).  This indicates a higher generalisation ability on the part of the spiking networks, as fewer classifiers are required to cover the entire state space.  Generalisation in STCS occurs in two ways, (i) where a single classifier can compute different (accurate) predictions in different environmental substates, and (ii) where a single classifier can calculate different actions based on differing state input.  Both types of generalisation appear in the final population.

We also note that self-adaptive mutation rates are globally lower in the spiking case, demonstrating the context-sensitivity of the self-adaptation process  --- Fig.~\ref{Spike-TCS}(d) shows spiking values ranging between 0.22 and 0.44 and Fig.~\ref{Spike-TCS}(e) shows MLP values ranging from 0.48 to 0.33 (all significantly different, P$<$0.05).  Lower mutation rates are indicitive of a more stable evolutionary process, as higher rates tend to be associated with a non-converged system.

Next, we decrease the step size from 0.05 to 0.005 and increase $t_{drop}$ from 20 to 200, preserving the average optimal macro steps to solve at 1.5.  All other parameters are unchanged.  We aim to (i) show that STCS is scaleable to larger state spaces that are more representative of those experienced in robotics problems, and (ii) demonstrate that the benefits of coupling a temporal RL scheme with a temporally sensitive neural classifier representation are enhanced as the amount of temporal information in the environment increases.

\begin{figure*}[t!]
\begin{center}

\subfloat[]{ \psfig{file=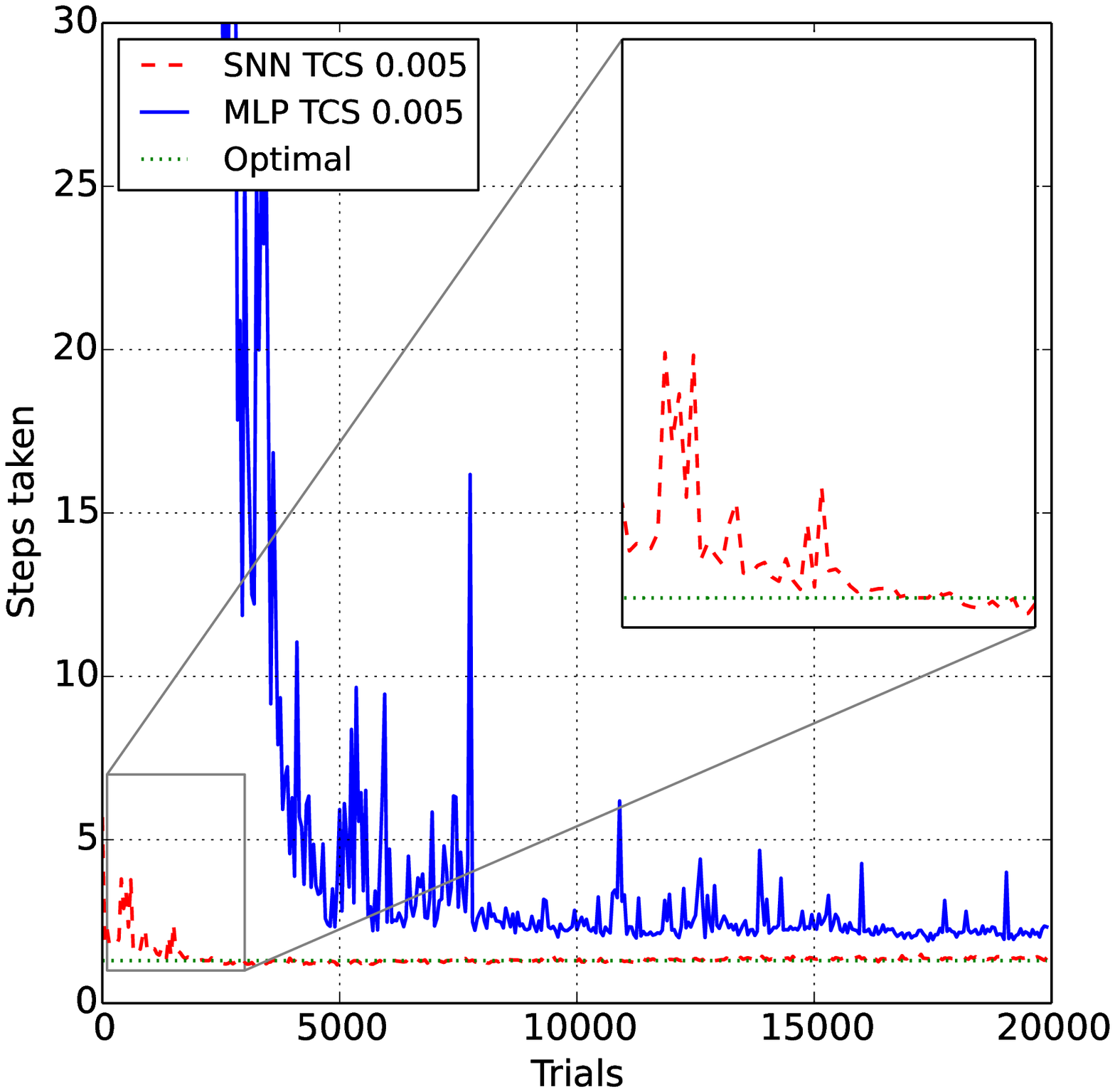,width=10cm,height=5cm}}\\
\subfloat[]{ \psfig{file=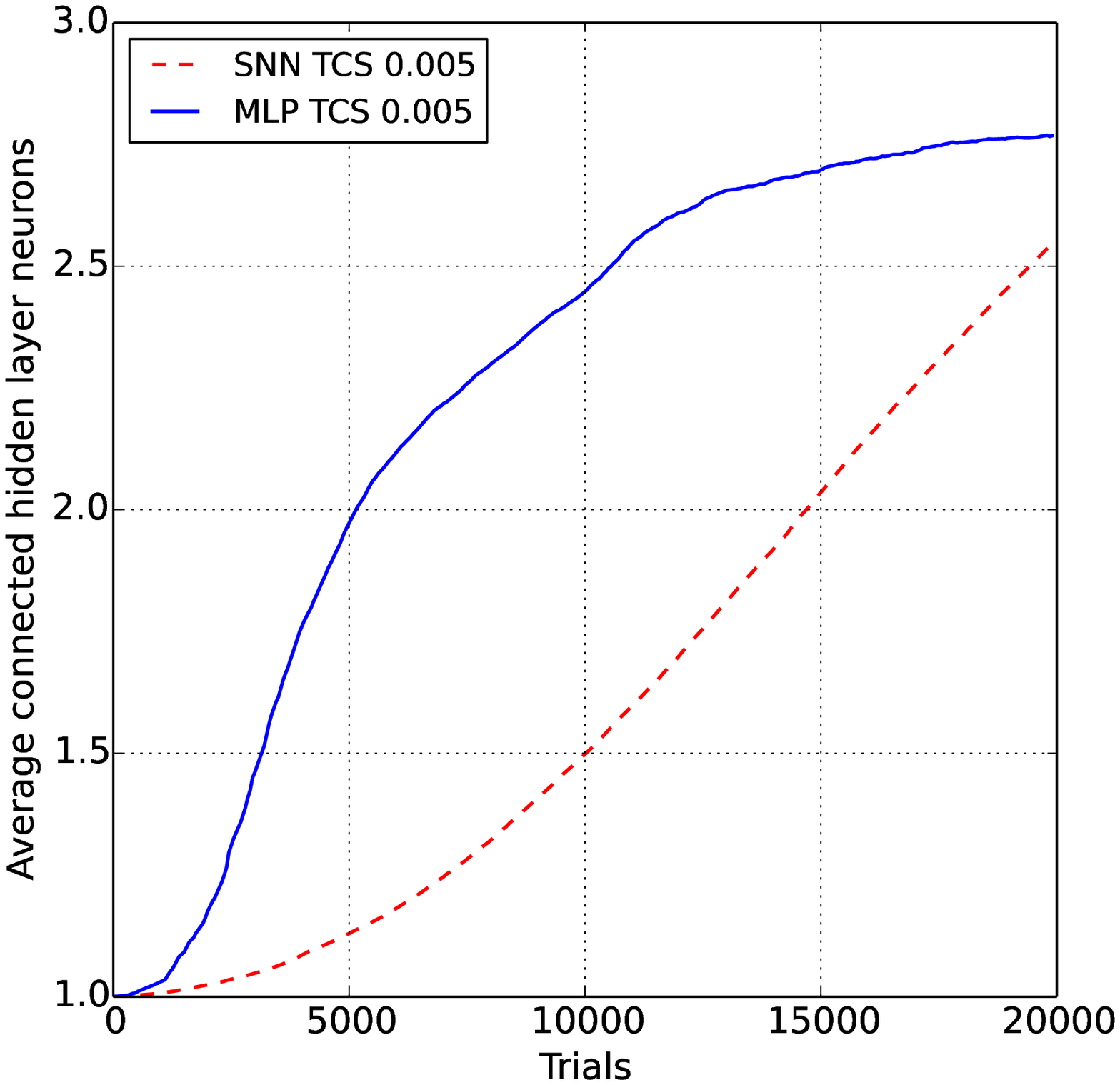,width=6cm,height=4cm}}
\subfloat[]{ \psfig{file=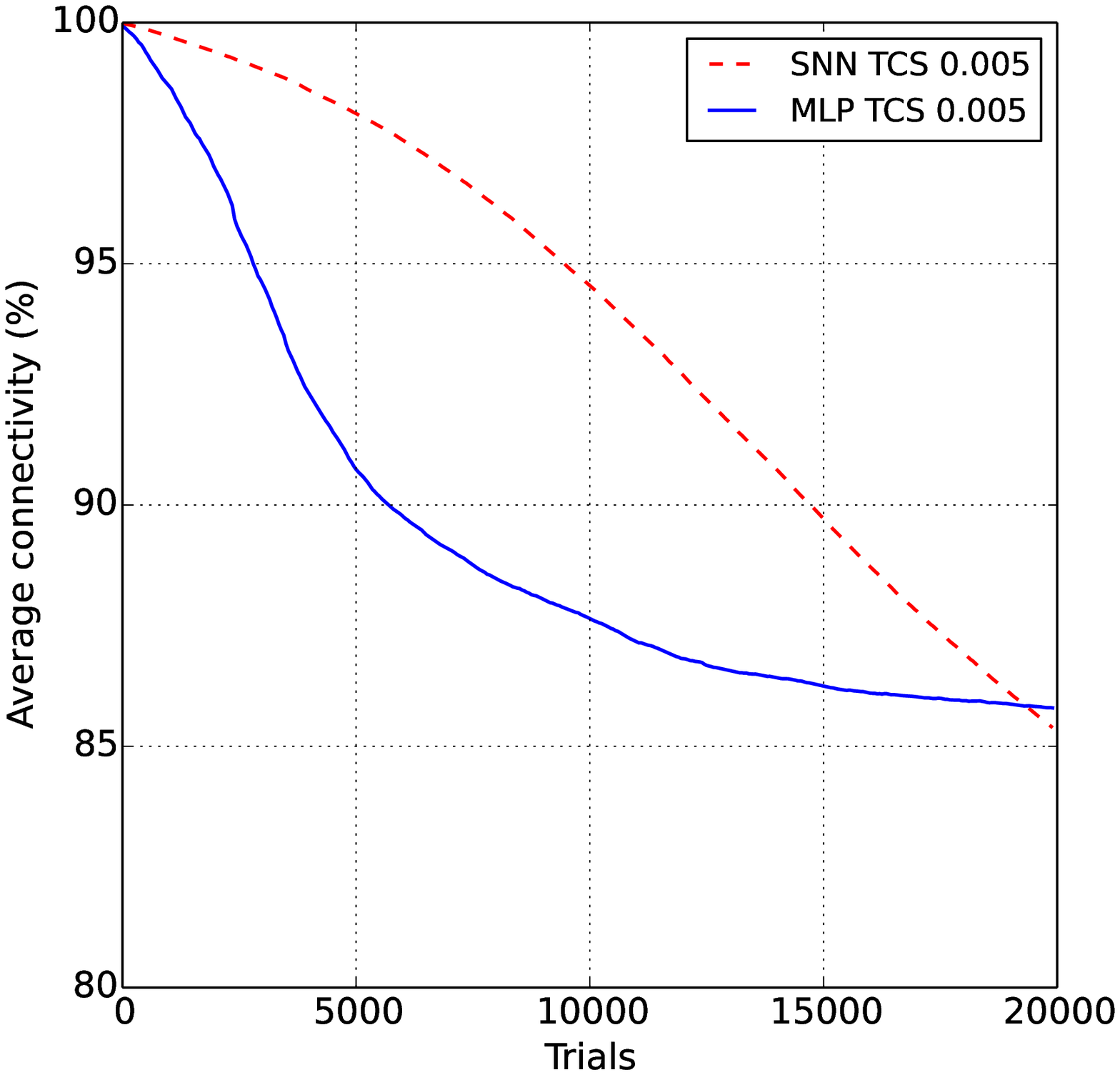,width=6cm,height=4cm}}\\
\subfloat[]{ \psfig{file=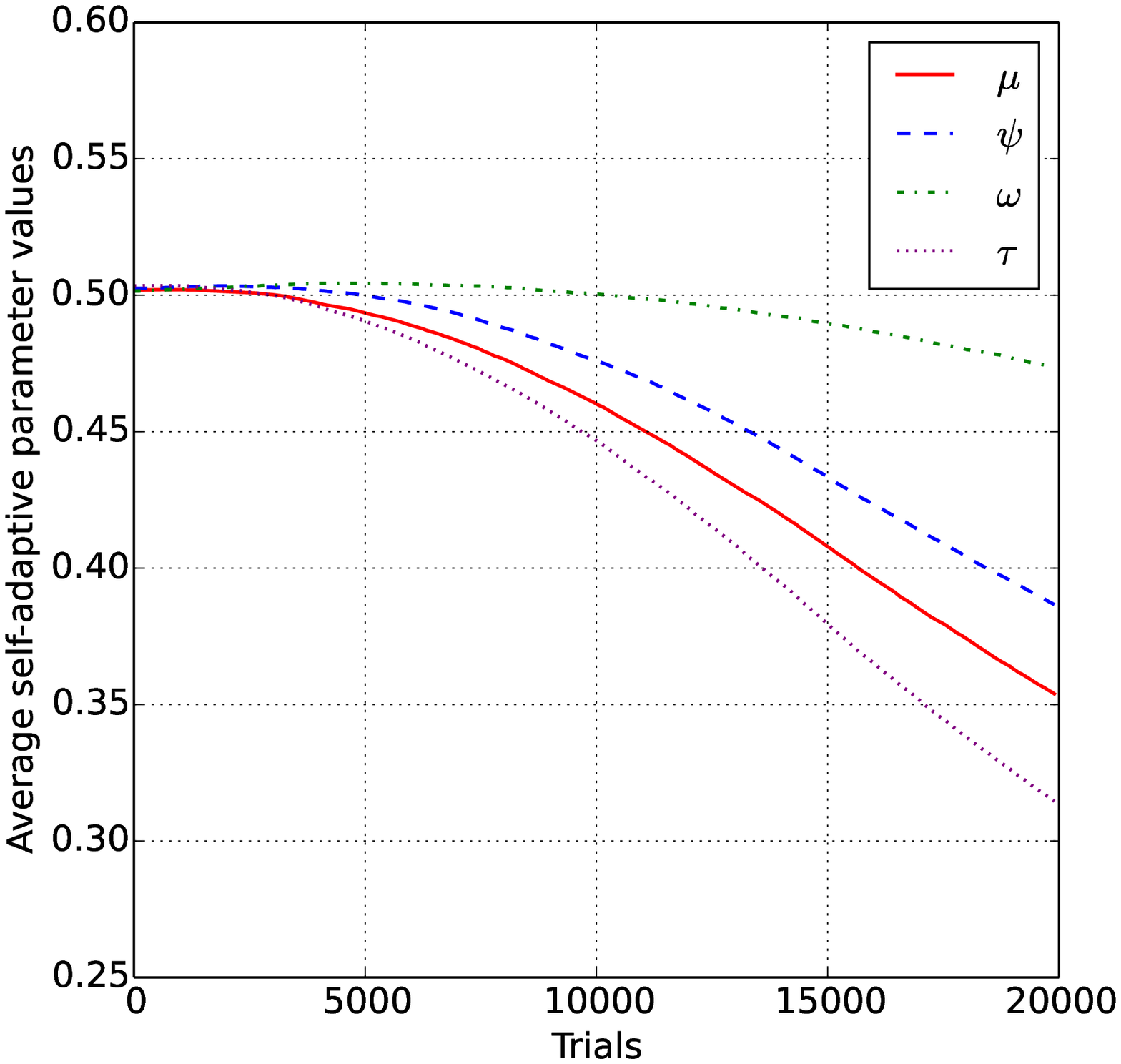,width=6cm,height=4cm}}
\subfloat[]{ \psfig{file=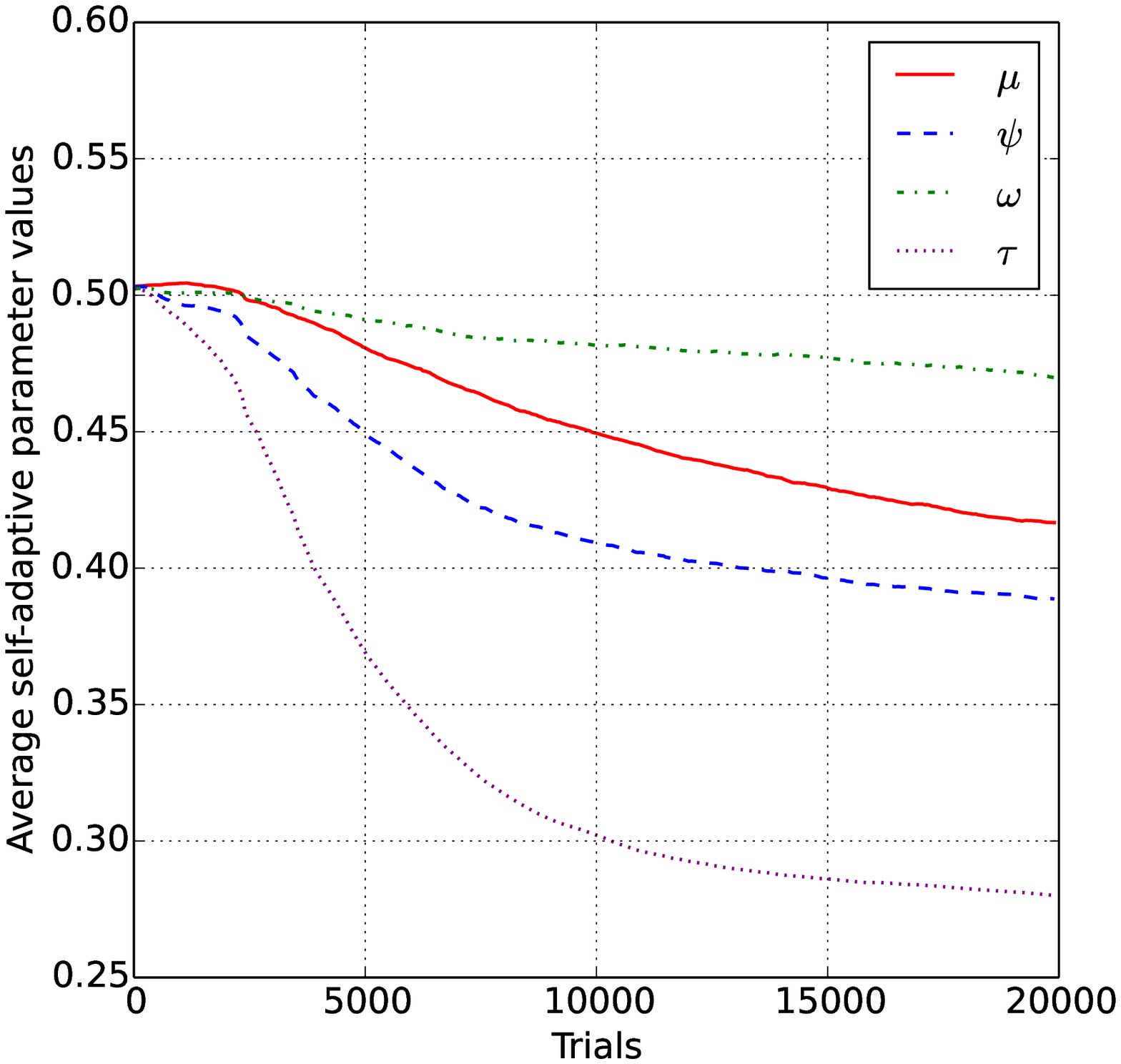,width=6cm,height=4cm}}

\end{center}
\caption[]{Continuous grid world 0.005 (a) average macro steps, (b) average connected hidden layer nodes, (c) average enabled connections, (d) average STCS self-adaptive parameter values, (e) average MLP TCS self-adaptive parameter values.}
\label{Spike-TCS-0005}
\end{figure*}

Fig.~\ref{Spike-TCS-0005}(a) shows that only STCS can achieve near-optimal performance (average macro steps 1.35, vs 1.48 for the MLP TCS, P$<$0.05).  MLP TCS never reaches a near-optimal steps value.  Analysis reveals that there are certian regions of the state space in which the classifiers in [A] cannot recalculate the action required to solve the problem.  MLP TCS must therefore drop the current action set and reform to successfully change action, taking a 1 macro step penalty when it is forced to do so.  As the MLP TCS cannot achieve optimal performance, it never achieves stability, whereas STCS does so in an average of 1242 trials (P$<$0.05).  This result is particularly important when use on robotic problems is considered --- many robotics problems will, in some regions of their state space, require highly heterogeneous actions where more complex behavioural policies, such as obstacle avoidance, are required.  As evidenced by the final steps values of STCS and MLP TCS, the spiking networks are more predisposed to allow for actions to be switched between without reforming a match set, as opposed to the more homogenous action selection evidenced in MLP networks (e.g., \cite{howard-gecco09}).

STCS is significantly better than MLP TCS with regards to macro steps required and stability.  As the main difference between the two representations is the temporal neural processing of the spiking networks, we conclude that the combination of temporal RL and a temporal classifier representation is beneficial to the performance of the system.  We note that the performance of the system increases as the amount of temporal information in the problem increases (181.4 vs. 18.4 average state transtions per trail in the 0.005/0.05 enviroment respectively).

STCS uses statictically fewer neurons than MLP TCS (2.51 vs 2.76, P$<$0.05), and is less connected (82.98\% vs 85.8\%, not statistically significant).  These results indicate that smaller networks can perform more complex state-action mappings thanks to the temporal inforamtion processing capabilities of the spiking networks.

Self-adaptive parameters are again universally lower in for STCS when compared to MLP TCS (0.416 vs 0.425 for $\mu$, 0.389 vs 0.412 for $\psi$, 0.47 vs 0.475 for $\omega$, and 0.280 vs 0.332 for $\tau$, P$<$0.05), which indicates a more stable evolutionary process.

\subsubsection{Comparing Identical Systems in the Different Environments}

A final analysis in the grid world compared STCS in the 0.05 grid world vs the 0.005 grid world --- Table~\ref{spike-TCS-005-0005}.  STCS solves both tasks optimally, however stability varies significantly between the two environments (avg 5637 for 0.05, avg 1244 for 0.005, P$<$0.05). A possible explanation for this is that, as the average number of discrete movements an agent is required to make is much greater in the 0.005 environment than in the step size 0.05 case, the spiking networks have more opportunity to use the temporal information in this semi-MDP, resulting in a performance difference.

STCS self-adaptive parameters Figs.~\ref{Spike-TCS}(d) and~\ref{Spike-TCS-0005}(d) in the two grid world environments follow an identical descending order ($\omega$, $\mu$, $\psi$, $\tau$) – the only difference being that the parameter values in Fig.~\ref{Spike-TCS}(c) are lower in general.  Final average self-adaptive mutation values vary statistically significantly between the different step size environments (all p$<$0.05).  Differing values again highlight the self-adaptive nature of the learning process, which alters depending on the environment the system is presented with.

We also note that MLP TCS performance degrades with increasing step size, whereas STCS improves.  In other words, STCS seems more suited to solving an environment with large numbers of state transitions, whereas MLPs lack the neural processing abilities required to handle such environments.

\begin{table}[t!]
\begin{center}
\caption[]{Detailing main averages and p-values results in STCS for the grid world with step sizes 0.05 and 0.005}
\label{spike-TCS-005-0005}
\begin{tabular}{|l|l|l|l|} 
\hline Metric & Step size & Average & P-value \\ 
\hline Stability & 0.05 & 5637.5 &  \\ 
  & 0.005 & 1244.13 & 2.42$\times$10$^{-6}$ \\ 
\hline Neurons & 0.05& 2.40 & \\
 & 0.005& 2.55&0.15 \\
\hline Connectivity &0.05& 86.13 & \\
& 0.005 &85.09 & 0.36 \\
\hline Macroclassifiers&0.05&16552.25 &  \\
 & 0.005 &18384.13 &0.03  \\
\hline 
\end{tabular}
\end{center}
\end{table}

\subsection {Robotics Problem}

\begin{figure}[t]
\begin{center}

\subfloat[]{ \psfig{file=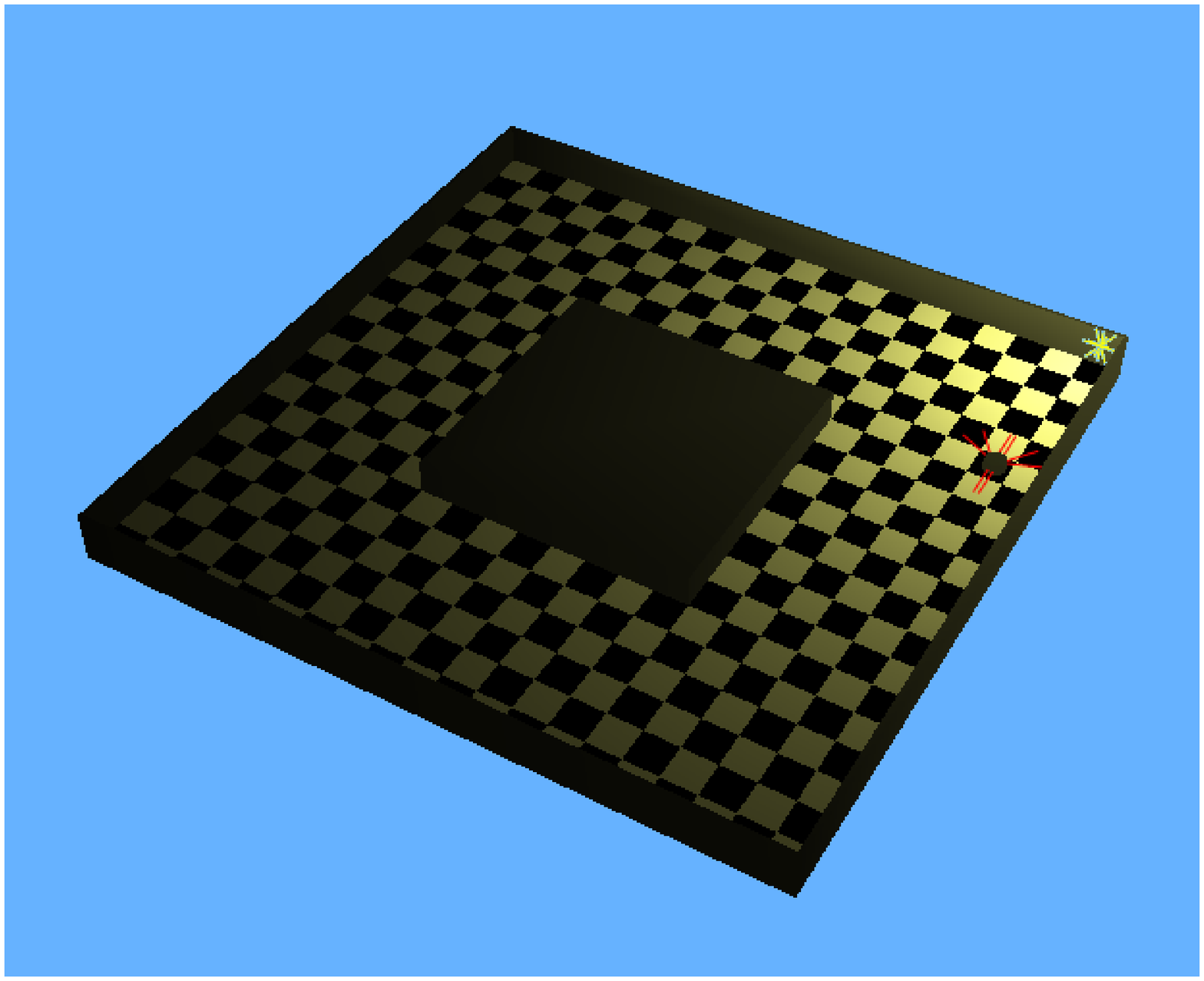,width=4cm, height=4cm}}
\subfloat[]{ \psfig{file=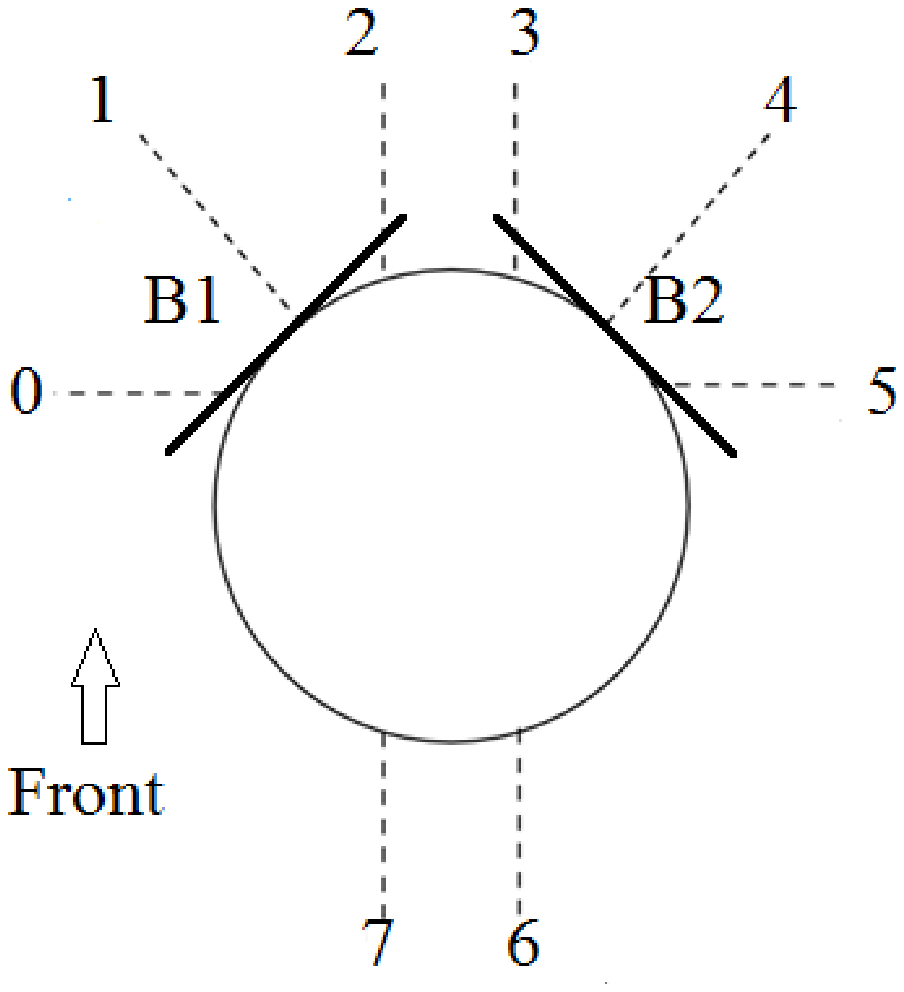,width=4cm, height=4cm}}
\subfloat[]{ \psfig{file=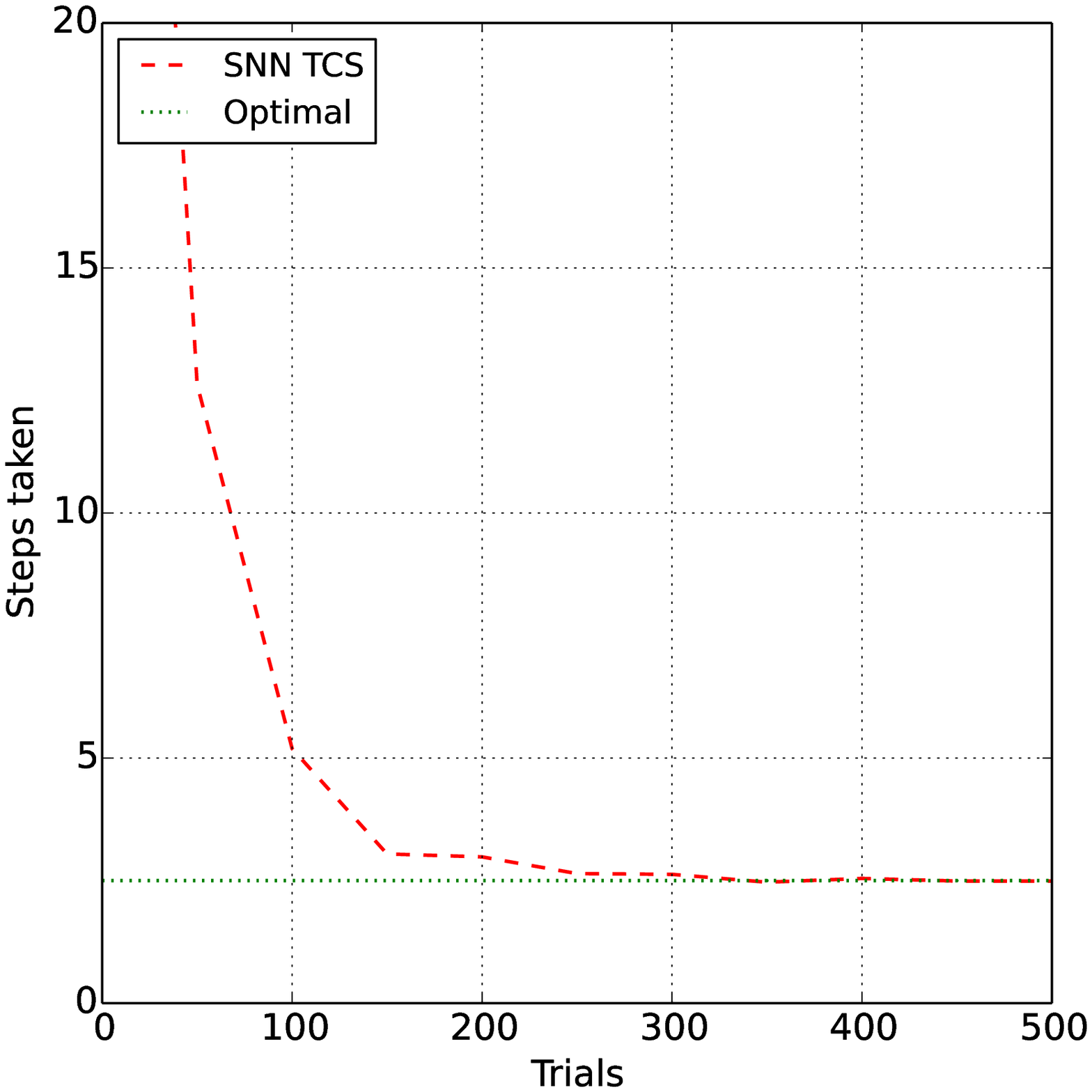,width=4cm,height=4cm}}

\end{center}
\caption[]{(a)The test environment.  The agent begins in the lower-left and must reach a light source in the upper-right, circumnavigating the central obstacle.(b)Khepera sensory arrangement.  3 light sensors and 3 IR sensors share positions 0, 2, and 5.  Two bump sensors, B1 and B2, are shown attached at 45 degree angles to the front-left and front-right of the robot. (c) Average macro steps for the experiment.}
\label{khep-sens}
\end{figure}

As a final evaluation, we test STCS on a robotics problem in which a agent must navigate towards a light source whilst avoiding an obstacle.  Our chosen robotics simulator is Webots ~\cite{webots04}.  The environment is a walled arena with coordinates ranging from [-1,1] in both $x$ and $y$ directions.  A box is placed centrally in the arena and a light source (modelled on a 15W bulb with realistic attenuation values) is placed at $x$=1, $y$=1, $z$=1. The agent's random start position is constrained (initial $x+y$ $<$ -1.5), forcing the agent to learn obstacle avoidance behaviour.  The environment is shown in Fig.~\ref{khep-sens}(a).

STCS commands a differential-drive agent with 3 light sensors and 3 distance sensors shown at positions 0, 2, and 5 in Fig.~\ref{khep-sens}(b) (all other IR/light sensors are disabled).  Random-uniform sensory noise in included -- $\pm$2\% for IR sensors and $\pm$10\% for light sensors.   In early trials, it is likely that the agent will bump into obstacles (before it has learned how to correctly respond to IR sensor readings).  Two bump sensors are placed to the front-left and front-right of the agent --- see Fig.~\ref{khep-sens}(b) --- activating a bumper causes the agent to drop the current action,  immediately reverse 10cm, and reform [M] (an effective penalty of 1 step).

 For each agent step (64ms in simulation time), the agent samples the six sensors: the six-dimensional input vector is then scaled so that the entire sensor range falls within [0,1], and is used as network input $I$ in equation \ref{eq3}.  Three actions are possible;  {\em forward}, (maximum movement on both wheels, {\em high} activation of the first two output neurons) and continuous turns to both the {\em left} ({\em high} activation on the first output neuron, {\em low} on the second) and {\em right} ({\em low} activation on the first output neuron, {\em high} on the second) --- caused by halving the left/right motor outputs respectively.  When the agent reaches the reward zone (where $x+y>$1.6), an immediate reward of 1000 is returned and the next trial begun.  All other movements give an immediate reward of 0.

This problem is chosen due to its difficulty --- the obstacle forces STCS to switch actions multiple times to achieve optimal performance.  The agent experiences realistic sensory noise and wheel slip.  The orientation of the agent is also taken into account, meaning the chaining together of individual actions into macro actions is more complex and dependent i.e, on the orientation of the agent with respect to the light source.  The input state is three times as large as in the grid world experiments, and the state space significantly larger (average optimal 640 state transitions per trial, compared to 181 for the 0.005 grid world and 18.4 for the 0.05 grid world).

Parameters are identical to previous experiments, except $N$=3000, $t_{drop}$=300, and each experiment lasts for 500 trials.  To allow the LCS to immediately make useful partitions in the state space, networks are intitially seeded with 6 hidden layer neurons and are initially 50\% connected.  This experiment has an optimal macro steps value of 2.5.

\subsection{Results}

Fig.~\ref{khep-sens}(c) shows the average macro step values attained.  Starting from 75 macro steps, the system shows swift attainment of near optimal performance  per trial after 300 trials.  Performance compares favourably to similar TCS robotics experiments, (e.g. \cite{journals/alife/HurstB06}).  Neurons and connectivity are largely unaltered from their initial values, due to the reduction in the number of trials.

\begin{figure*}[ht!]
\begin{center}

\subfloat[]{ \psfig{file=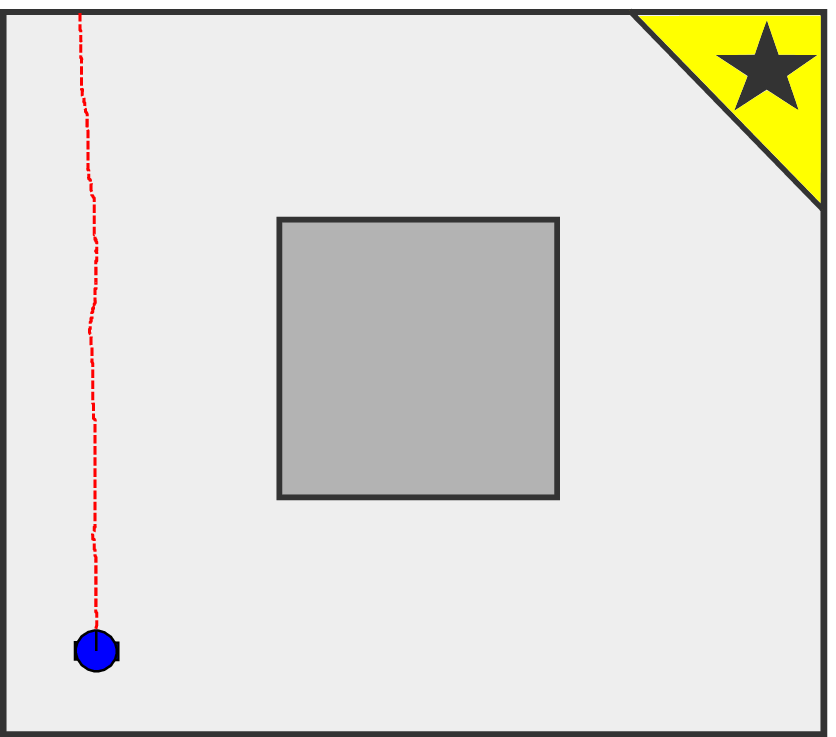,width=4cm,height=4cm}}
\subfloat[]{ \psfig{file=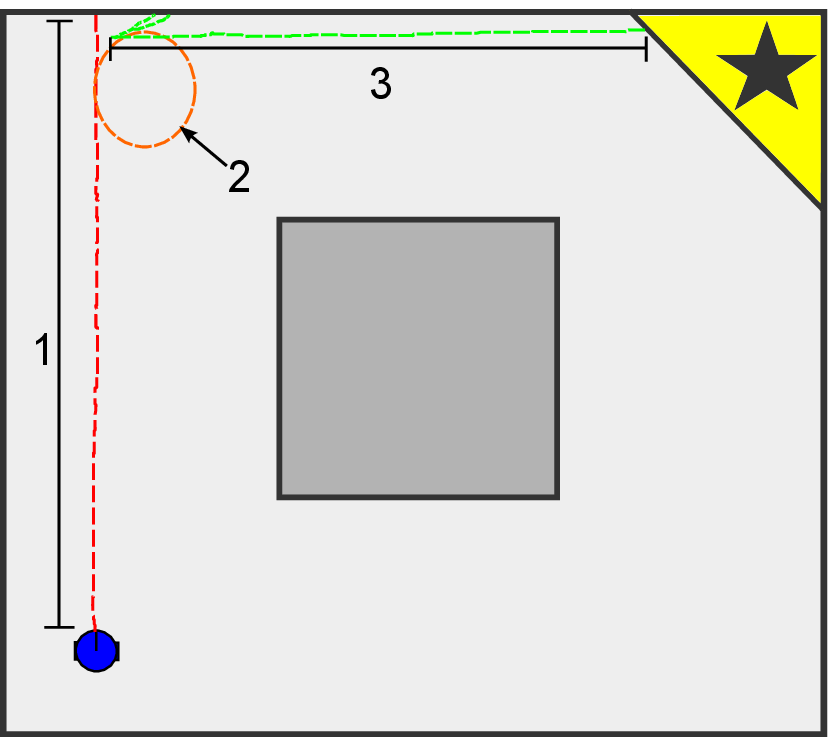,width=4cm,height=4cm}}
\subfloat[]{ \psfig{file=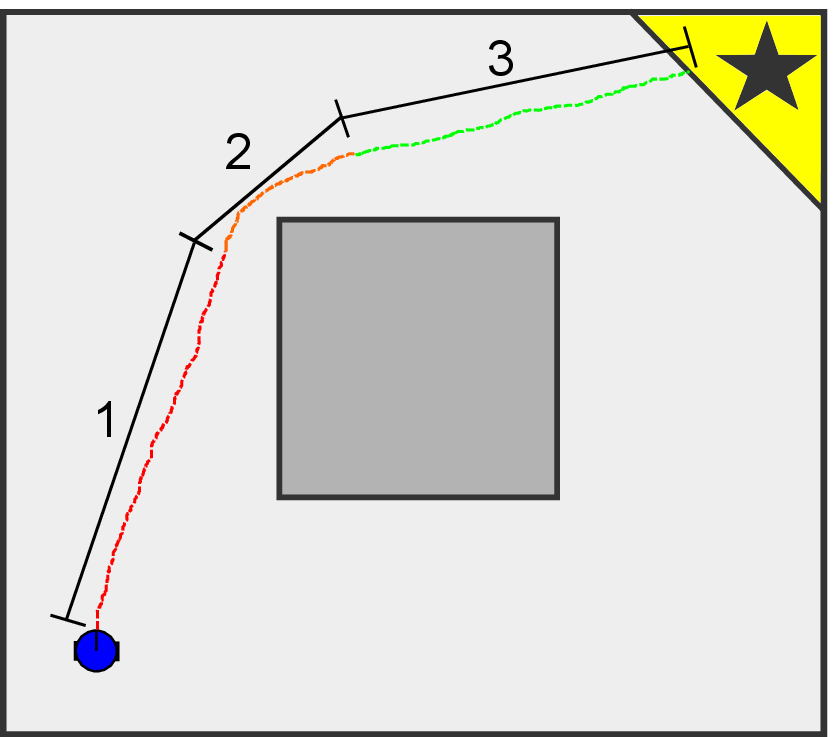,width=4cm,height=4cm}}

\end{center}
\caption[]{Showing typical agent paths at various stages of the experiment (goal state denoted with a star). (a) Initially, the agent repeatedly collides with a wall until the trial step limit is reached. (b) After approx. 50 trials, the agent initially reaches the goal state by (1) colliding with a wall, (2) turning until the step limit is reached, then repeatedly colliding with the wall until the robot aligns with the goal state, (3) finally proceeding to the goal state. (c) Finally, the agent learns to navigate to the goal state in a minimal number of steps (initially 3 match sets are formed, which can then be simplified into fewer match sets through action recalculation).
}
\label{webots-paths}
\end{figure*}

Fig.\ref{webots-paths} shows how the generated path quality in exploit mode improves through successive trials.  Initially, the agent bumps into walls repeatedly and fails to find the goal state.  After approximately 50 trials, the agent finds the goal state (although this takes many [M] formations).  The final image shows a sample path after 150 generations, requiring three [M] formations (labelled) but using action switching within those sets.  Continued improvements increase the duration of each macro action, reducing the number of macro actions required on average.

STCS can --- in most cases --- evolve networks that perform segregations in action space without dropping [A] and reforming [M] via the majority of the networks in [A] recalculating their actions to change the action advocated by [A] in response to some state input.  Action alteration is usually due to IR sensors perturbing the networks action to turn the agent away from an obstacle or towards the light source, although the agent is also observed to display the capability to alter action based on light sensor input alone.  A lesser-used action switching mechanism involved networks activating the ``don’t match'' node so that networks advocating certain actions are dropped from [A] at the correct time.  This made it more likely for the system action to switch.Considering all in-[A] action switches, the first method is used significantly more frequently (P$<$0.05) as it kept more classifiers in [A], helping to retain classifier variety for future switches.  Subsequent trials --- Fig.\ref{webots-paths}(c) --- display increased amounts of action switching, which reduces the average number of [M] formations required per trial to the optimum value whilst preserving the overall path quality.

\section{Conclusions and Outlook}

In this article we have presented a cognitive architecture based on temporal reinforcement learning and spiking networks.  Our hypothesis --- that the coupling of temporal RL (from TCS) and neural processing (from the spiking networks) is beneficial to the system, has been shown throughout our results.  STCS allows the generation --- and accurate payoff assignment --- of high-level macro actions from heterogenous chains of simple discrete actions, which allows the agent to handle environments requiring large amounts of state transitions.  Unlike many RL approaches to solve continuous-state spaces, STCS can solve semi-MDP environments without having to prediscretise the state space.  Our results show that:

\begin{itemize}

\item STCS uses fewer macroclassifiers than MLP TCS on the grid world 0.05 environment, indicating more generalisation ability.
\item STCS attains significantly better performance than MLP TCS in the grid world 0.005, where the main difference is the provision of neural processing in STCS.  Latent temporal information in the problem is effectively harnessed by the spiking networks, resulting in this performance gain.
\item STCS increases performance with increasing numbers of required state transitions, whereas MLP TCS performance degrades.
\item STCS permits simpler temporal discretisations in large, complex state spaces than MLP TCS.  Spiking classifiers aid in allowing the recalculation of actions based on both the current state input and, uniquely, the networks own internal states.

\end{itemize}

The final point is particularly important, as it highlights the complimentary nature of the temporal RL/spiking network combination --- by providing the classifiers with more computational power (i.e., neural processing, as is not done in other TCS approaches), the system can learn the simplest, optimal discretisations of state space into required behaviours, potentially comprising long, complex chains of actions through action recalculation.

As a prototypical cognitive architecture, we also wish to highlight the self-adaptive elements of STCS.  Self-adaptation is used to achieve a level of parameter independence that is rarely seen in Learning Classifier Systems.  Both LCS and RL can be sensitive to various parameter selections.  Throughout all experiments, the only main parameters changed are $t_{drop}$, population size, and the number of trials.  Self-adaptation largely precludes the need for parameter sweeps, which are required in other TCS systems (Hurst note strong parameter depedence in some of their TCS research \cite{Hurst:2002:TLC}).  Self-adaptive parameters vary significantly between the individual rates, the systems they are implemented in (STCS/MLP TCS),  and the problems those systems solve, highlighting their context-sensitivity.

When self-adapting spiking network parameters, the effect is that the evolutionary process itself --- and as a result, the networks it produces --- is tailored to features of the environment.  Constructive neuro-evolution is shown to generate context-sensitive structures that encode important features of robotic navigation, which can be specific to the environmental states they are active in.  By starting with simple neural structures, the evolutionary process itself is expediated and approximately minimally-complex networks (in terms of neurons and connections) that encode useful {\em state, action} combinations are discovered in the regions of the state space in which they are selected.

Since classifier prediction, classifier action, and classifier matching are all computed, generality is promoted as a single classifier can (i) match all regions of the state space, (ii) advocate different actions in different state subspaces, and (iii) accurately predict the payoffs of all state, action pairs which it encodes.   We note that it is theoretically possible for a single classifier to solve entire problems, by calculating the best action and accurate predictions in every state.  This does not occur as it is much simpler for the LCS to evolve ensembles of simpler, specialised networks.

We believe the best direction to take this work is into the realm of increased biological realism, i.e., replacing some of the more mechanistic aspects of LCS (set formations, etc.) with counterparts that are more rigourously grounded in computational neuroscience.  Biologically-realistic extensions can also be applied to the spiking networks, which are particularly amenable to a form of unsupervised learning known as spike-time-dependent plasticity (STDP) \cite{stdp}.  This would more closely couple the neural processing to features of the environment, by allowing a network to react to subsequent input states by varying its connection weights during a trial.  This has been discussed, but not implemented, in the context of neural replication by Fernando \cite{replicators}.

\bibliographystyle{spbasic}
\bibliography{xcsf-sc}

\begin{thebibliography}{52}
\providecommand{\natexlab}[1]{#1}
\providecommand{\url}[1]{{#1}}
\providecommand{\urlprefix}{URL }
\expandafter\ifx\csname urlstyle\endcsname\relax
  \providecommand{\doi}[1]{DOI~\discretionary{}{}{}#1}\else
  \providecommand{\doi}{DOI~\discretionary{}{}{}\begingroup
  \urlstyle{rm}\Url}\fi
\providecommand{\eprint}[2][]{\url{#2}}

\bibitem[{Beer(1995)}]{beer}
Beer RD (1995) On the dynamics of small continuous-time recurrent neural
  networks. Adapt Behav 3:469--509, \doi{10.1177/105971239500300405},
  \urlprefix\url{http://portal.acm.org/citation.cfm?id=218530.218539}

\bibitem[{Bonarini(1998)}]{Bonarini1998a}
Bonarini A (1998) Reinforcement distribution to fuzzy classifiers. In:
  {Proceedings of the IEEE World Congress on Computational Intelligence (WCCI)
  -- Evolutionary Computation}, {IEEE Computer Press}, pp 51--56

\bibitem[{Boyan and Moore(1995)}]{boyan.moore-1995:gener}
Boyan JA, Moore AW (1995) Generalization in reinforcement learning: {S}afely
  approximating the value function. In: Tesauro G, Touretzky DS, Leen TK (eds)
  Advances in Neural Information Processing Systems 7, The MIT Press,
  Cambridge, MA, pp 369--376

\bibitem[{Bull(2002)}]{Bull:2002:UCN}
Bull L (2002) On using constructivism in neural classifier systems. In: Merelo
  J, Adamidis P, Beyer HG, Fernandez-Villacanas JL, Schwefel HP (eds) Parallel
  Problem Solving from Nature - PPSN VII, Springer Verlag, pp 558--567

\bibitem[{Bull and Hurst(2003)}]{bull-hurst-tech03}
Bull L, Hurst J (2003) A neural learning classifier system with self-adaptive
  constructivism. In: Proceedings of the IEEE Congress on Evolutionary
  Computation, IEEE Press, pp 991--997

\bibitem[{Butz and Herbort(2008)}]{conf/gecco/ButzH08}
Butz MV, Herbort O (2008) Context-dependent predictions and cognitive arm
  control with {XCSF}. In: Ryan C, Keijzer M (eds) Genetic and Evolutionary
  Computation Conference, {GECCO} 2008, Proceedings, Atlanta, {GA}, {USA}, July
  12-16, 2008, ACM, pp 1357--1364

\bibitem[{Butz et~al(2006)Butz, Lanzi, and Wilson}]{lanzi:2006:hyp}
Butz MV, Lanzi PL, Wilson SW (2006) Hyper-ellipsoidal conditions in xcs:
  rotation, linear approximation, and solution structure. In: GECCO '06:
  Proceedings of the 8th annual conference on Genetic and evolutionary
  computation, ACM Press, New York, NY, USA, pp 1457--1464,
  \doi{http://doi.acm.org/10.1145/1143997.1144237}

\bibitem[{Buzsaki(2006)}]{rhythms}
Buzsaki G (2006) Rhythms of the Brain. Oxford University Press

\bibitem[{Cazangi et~al(2003)Cazangi, Zuben, and
  Figueiredo}]{conf/cec/CazangiZF03}
Cazangi RR, Zuben FJV, Figueiredo M (2003) A classifier system in real
  applications for robot navigation. In: IEEE Congress on Evolutionary
  Computation, IEEE, pp 574--580

\bibitem[{Churchill and Fernando(2014)}]{churchill2014evolutionary}
Churchill AW, Fernando C (2014) An evolutionary cognitive architecture made of
  a bag of networks. Evolutionary Intelligence 7(3):169--182

\bibitem[{Donnart and Meyer(1996)}]{Donnart1996c}
Donnart JY, Meyer JA (1996) Learning reactive and planning rules in a
  motivationally autonomous animat. IEEE Transactions on Systems, Man and
  Cybernetics - Part B: Cybernetics 26(3):381--395

\bibitem[{Dorigo and Colombetti(1994)}]{journals/ai/DorigoC94}
Dorigo M, Colombetti M (1994) Robot shaping: Developing autonomous agents
  through learning. Artificial Intelligence 71(2):321--370

\bibitem[{Faußer and Schwenker(2015)}]{nn-ensemble-rl}
Faußer S, Schwenker F (2015) Neural network ensembles in reinforcement
  learning. Neural Processing Letters 41(1):55--69,
  \doi{10.1007/s11063-013-9334-5}

\bibitem[{Fernando(2011)}]{fernandoLCS}
Fernando C (2011) Symbol manipulation and rule learning in spiking neuronal
  networks. Journal of theoretical biology 275(1):29--41

\bibitem[{Fernando et~al(2010)Fernando, Goldstein, and
  Szathm{\'a}ry}]{replicators}
Fernando C, Goldstein R, Szathm{\'a}ry E (2010) The neuronal replicator
  hypothesis. Neural Computation 22(11):2809--2857

\bibitem[{Floreano and Mattiussi(2001)}]{Floreano:2001:ESN}
Floreano D, Mattiussi C (2001) Evolution of spiking neural controllers for
  autonomous vision-based robots. Lecture Notes in Computer Science 2217:38--61

\bibitem[{Floreano et~al(2002)Floreano, Schoeni, Caprari, and Blynel}]{evo-bns}
Floreano D, Schoeni N, Caprari G, Blynel J (2002) Evolutionary
  bits’n’spikes. In: In Artificial Life VIII Proceedings, MIT Press

\bibitem[{Gerstner and Kistler(2002)}]{spiking-n-m}
Gerstner W, Kistler W (2002) Spiking Neuron Models - Single Neurons,
  Populations, Plasticity. Cambridge University Press

\bibitem[{Hagras and Sobh(2002)}]{hagras2002intelligent}
Hagras H, Sobh T (2002) Intelligent learning and control of autonomous robotic
  agents operating in unstructured environments. Information Sciences
  145(1):1--12

\bibitem[{He and Jagannathan(2007)}]{he2007reinforcement}
He P, Jagannathan S (2007) Reinforcement learning neural-network-based
  controller for nonlinear discrete-time systems with input constraints.
  Systems, Man, and Cybernetics, Part B: Cybernetics, IEEE Transactions on
  37(2):425--436

\bibitem[{Hodgkin and Huxley(1952)}]{hodgkin-huxley}
Hodgkin AL, Huxley AF (1952) A quantitative description of membrane current and
  its application to conduction and excitation in nerve. The Journal of
  physiology 117(4):500

\bibitem[{Holland(1975)}]{Holland75}
Holland JH (1975) Adaptation in Natural and Artificial Systems. The University
  of Michigan Press, Ann Arbor, Michigan

\bibitem[{Holland(1976)}]{Holland76}
Holland JH (1976) Adaptation. In: Rosen R, Snell F (eds) Progress in
  Theoretical Biology, Academic Press

\bibitem[{Holland and Reitman(1978)}]{Holland1978}
Holland JH, Reitman JS (1978) Cognitive systems based on adaptive algorithms.
  In: Waterman DA, Hayes-Roth F (eds) Pattern-Directed Inference Systems,
  Academic Press, Orlando, pp 313--329

\bibitem[{Howard et~al(2010)Howard, Bull, and Lanzi}]{howard2010spiking}
Howard G, Bull L, Lanzi PL (2010) A spiking neural representation for xcsf. In:
  IEEE Congress on Evolutionary Computation (CEC), IEEE, pp 1--8

\bibitem[{Howard and Bull(2008)}]{conf/gecco/HowardB08}
Howard GD, Bull L (2008) On the effects of node duplication and
  connection-oriented constructivism in neural {XCSF}. In: Ryan C, Keijzer M
  (eds) Genetic and Evolutionary Computation Conference, {GECCO} 2008,
  Proceedings, Atlanta, {GA}, {USA}, July 12-16, 2008, Companion Material, ACM,
  pp 1977--1984

\bibitem[{Howard et~al(2009)Howard, Bull, and Lanzi}]{howard-gecco09}
Howard GD, Bull L, Lanzi PL (2009) Towards continuous actions in continuous
  space and time using self-adaptive constructivism in neural {XCSF}. In: GECCO
  '09: Proceedings of the 11th Annual conference on Genetic and evolutionary
  computation, ACM, New York, NY, USA, pp 1219--1226,
  \doi{http://doi.acm.org/10.1145/1569901.1570065}

\bibitem[{Hurst and Bull(2006)}]{journals/alife/HurstB06}
Hurst J, Bull L (2006) A neural learning classifier system with self-adaptive
  constructivism for mobile robot control. Artificial Life 12(3):353--380

\bibitem[{Hurst et~al(2002)Hurst, Bull, and Melhuish}]{Hurst:2002:TLC}
Hurst J, Bull L, Melhuish C (2002) {TCS} learning classifier system controller
  on a real robot. Lecture Notes in Computer Science 2439:588--600

\bibitem[{Kistler(2002)}]{stdp}
Kistler WM (2002) Spike-timing dependent synaptic plasticity: a
  phenomenological framework. Biological Cybernetics 87(5-6):416--427,
  \urlprefix\url{http://dx.doi.org/10.1007/s00422-002-0359-5}

\bibitem[{Lanzi and Loiacono(2006)}]{xcsf-neuralpred}
Lanzi P, Loiacono D (2006) Xcsf with neural prediction. In: Yen GG, Lucas SM,
  Fogel G, Kendall G, Salomon R, Zhang BT, Coello CAC, Runarsson TP (eds)
  Proceedings of the 2006 IEEE Congress on Evolutionary Computation, IEEE
  Press, Vancouver, BC, Canada, pp 2270--2276,
  \urlprefix\url{http://ieeexplore.ieee.org/servlet/opac?punumber=11108}

\bibitem[{Lanzi et~al(2005)Lanzi, Loiacono, Wilson, and
  Goldberg}]{conf/cec/LanziLWG05a}
Lanzi PL, Loiacono D, Wilson SW, Goldberg DE (2005) {XCS} with computed
  prediction in continuous multistep environments. In: IEEE Congress on
  Evolutionary Computation, IEEE, pp 2032--2039

\bibitem[{Lanzi et~al(2006)Lanzi, Loiacono, Wilson, and Goldberg}]{lanzi-tile}
Lanzi PL, Loiacono D, Wilson SW, Goldberg DE (2006) Classifier prediction based
  on tile coding. In: Proceedings of the 8th annual conference on Genetic and
  evolutionary computation, ACM, New York, NY, USA, GECCO '06, pp 1497--1504

\bibitem[{Maass(1997)}]{maass}
Maass W (1997) Networks of spiking neurons: the third generation of neural
  network models. Neural networks

\bibitem[{Michel(2004)}]{webots04}
Michel O (2004) Webots{TM}: Professional mobile robot simulation. International
  Journal of Advanced Robotic Systems 1(1):39--42

\bibitem[{Moioli et~al(2007)Moioli, Vargas, and Zuben}]{conf/iwcls/MoioliVZ07}
Moioli RC, Vargas PA, Zuben FJV (2007) Analysing learning classifier systems in
  reactive and non-reactive robotic tasks. In: Bacardit J, Bernad{\'o}-Mansilla
  E, Butz MV, Kovacs T, Llor{\`a} X, Takadama K (eds) International Workshop on
  Learning Classifier Systems {IWLCS}, Springer, Lecture Notes in Computer
  Science, vol 4998, pp 286--305

\bibitem[{Pipe and Carse(2002)}]{Pipe:2002:FRE}
Pipe AG, Carse B (2002) First results from experiments in fuzzy classifier
  system architectures for mobile robotics. Lecture Notes in Computer Science
  2439:578--587

\bibitem[{Preen and Bull(2014)}]{DAFDGP}
Preen R, Bull L (2014) Discrete and fuzzy dynamical genetic programming in the
  xcsf learning classifier system. Soft Computing 18(1):153--167,
  \doi{10.1007/s00500-013-1044-4},
  \urlprefix\url{http://dx.doi.org/10.1007/s00500-013-1044-4}

\bibitem[{Quartz and Sejnowski(1997)}]{QandS}
Quartz SR, Sejnowski TJ (1997) The neural basis of cognitive development: A
  constructivist manifesto. Behavioral and Brain Sciences

\bibitem[{Rechenberg(1973)}]{rechenberg}
Rechenberg I (1973) {Evolutionsstrategie: optimierung technischer systeme nach
  prinzipien der biologischen evolution}. Frommann-Holzboog

\bibitem[{Rumelhart and McClelland(1986)}]{rumelhart86}
Rumelhart D, McClelland J (1986) Parallel Distributed Processing, vol 1 \& 2.
  MIT Press, Cambridge, MA

\bibitem[{Schultz(1998)}]{schultz}
Schultz W (1998) Predictive reward signal of dopamine neurons. Journal of
  Neurophysiology 80(1):1--27

\bibitem[{Shouval and Gavornik(2011)}]{single-snn-neuron}
Shouval H, Gavornik J (2011) A single spiking neuron that can represent
  interval timing: analysis, plasticity and multi-stability. Journal of
  Computational Neuroscience 30(2):489--499

\bibitem[{Stolzmann(1999)}]{stolzmann}
Stolzmann W (1999) Latent learning in khepera robots with anticipatory
  classifier systems. In: Lanzi PL, Stolzmann W, Wilson SW (eds) 2nd
  International Workshop on Learning Classifier Systems, Orlando, Florida, USA,
  pp 290--297

\bibitem[{Studley and Bull(2005)}]{conf/cec/StudleyB05}
Studley M, Bull L (2005) {X}-{TCS}: accuracy-based learning classifier system
  robotics. In: IEEE Congress on Evolutionary Computation, IEEE, pp 2099--2106

\bibitem[{Sutton(1996)}]{sutton-scc}
Sutton RS (1996) Generalization in reinforcement learning: Successful examples
  using sparse coarse coding. In: Advances in Neural Information Processing
  Systems 8, MIT Press, pp 1038--1044

\bibitem[{Sutton et~al(1999)Sutton, Precup, and Singh}]{sutton-99}
Sutton RS, Precup D, Singh S (1999) Between mdps and semi-mdps: a framework for
  temporal abstraction in reinforcement learning. Artificial Intelligence
  112(1-2):181--211

\bibitem[{Watkins(1989)}]{watkins-thesis}
Watkins C (1989) Learning from delayed rewards. PhD thesis, Cambridge
  University, Psychology Dept., Cambridge, UK

\bibitem[{Webb et~al(2003)Webb, Hart, Ross, and Lawson}]{conf/ecal/WebbHRL03}
Webb A, Hart E, Ross P, Lawson A (2003) Controlling a simulated khepera with an
  {XCS} classifier system with memory. In: Banzhaf W, Christaller T, Dittrich
  P, Kim JT, Ziegler J (eds) Advances in Artificial Life, 7th European
  Conference, {ECAL} 2003, Dortmund, Germany, September 14-17, 2003,
  Proceedings, Springer, Lecture Notes in Computer Science, vol 2801, pp
  885--892

\bibitem[{Wilson(2000)}]{Wilson1999b}
Wilson SW (2000) Get real! xcs with continuous-valued inputs. In: Learning
  Classifier Systems, From Foundations to Applications, LNAI-1813,
  Springer-Verlag, pp 209--219

\bibitem[{Wilson(2001{\natexlab{a}})}]{Wilson2001a}
Wilson SW (2001{\natexlab{a}}) Function approximation with a classifier system.
  In: Spector L, Goodman ED, Wu A, Langdon WB, Voigt HM, Gen M, Sen S, Dorigo
  M, Pezeshk S, Garzon MH, Burke E (eds) Proceedings of the Genetic and
  Evolutionary Computation Conference (GECCO-2001), Morgan Kaufmann, San
  Francisco, California, USA, pp 974--981

\bibitem[{Wilson(2001{\natexlab{b}})}]{Wilson2001b}
Wilson SW (2001{\natexlab{b}}) Mining oblique data with {XCS}. In: Lanzi PL,
  Stolzmann W, Wilson SW (eds) Advances in learning classifier systems, third
  international workshop, {IWLCS} 2000, LNCS, vol 1996, Springer, pp 158--176

\end{thebibliography}

\end{document}